%% file: main.tex
\documentclass[11pt,a4paper]{wan}

\usepackage{amsmath,amssymb}
\usepackage{booktabs}
\usepackage{makecell}
\usepackage{tabularx}
\usepackage{array}
\usepackage{enumitem}
\usepackage{xspace}
\usepackage{placeins}
\usepackage{float}
\usepackage{xcolor}
\usepackage{colortbl}
\usepackage{tikz}
\usetikzlibrary{arrows.meta,positioning,fit,calc,shapes.geometric}

\definecolor{wanPrimary}{RGB}{97,92,237}
\definecolor{wanPrimaryDark}{RGB}{72,68,196}
\definecolor{wanPrimarySoft}{RGB}{242,241,255}
\definecolor{wanBorder}{RGB}{219,217,246}
\definecolor{cbBlue}{RGB}{0,114,178}
\definecolor{cbOrange}{RGB}{230,159,0}
\definecolor{tableGroup}{rgb}{0.95,0.95,1.0}
\definecolor{tableBest}{rgb}{0.925,0.7,0.775}
\definecolor{tableSecond}{rgb}{0.9625,0.85,0.8875}
\newcommand{\BestScore}[1]{\cellcolor{tableBest}\textbf{\boldmath#1}}
\newcommand{\SecondScore}[1]{\cellcolor{tableSecond}#1}

\input{math_commands.tex}

\usepackage{hyperref}
\hypersetup{
  pdftitle={DuplexBridge: Coordinating Understanding and Generation for Unified Image Editing},
  pdfauthor={Yidi Liu, Jie Xiao, Chaojie Mao, Jingfeng Zhang, Xueyang Fu},
  pdfsubject={DuplexBridge arXiv Technical Report}}
\usepackage[noabbrev,nameinlink]{cleveref}
\usepackage{url}

\input{results_current/numbers}

\input{results_added/numbers}
\title{DuplexBridge: Coordinating Understanding\\
and Generation for Unified Image Editing}

\author[1]{Yidi Liu}
\author[2]{Jie Xiao}
\author[2]{Chaojie Mao}
\author[2]{Jingfeng Zhang}
\author[1,\ast]{Xueyang Fu}
\affiliation[1]{University of Science and Technology of China\\}
\affiliation[2]{Wan Team, Alibaba Group}
\contribution[\ast]{Corresponding author}
\date{October 12, 2026}
\abstract{\input{sections/0_abstract}}

\fancypagestyle{firstheader}{%
  \fancyhf{}
  \fancyhead[R]{\includegraphics[height=0.64cm]{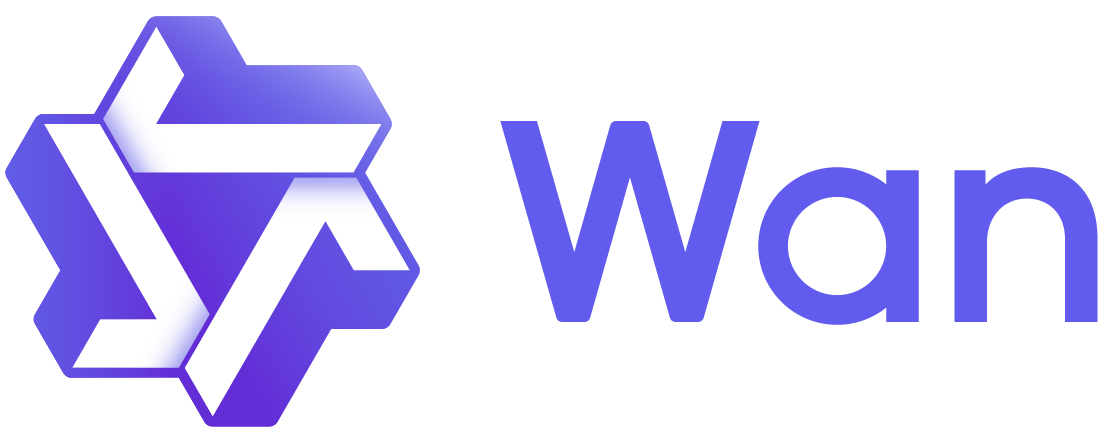}}
  \fancyfoot[C]{\small\thepage}
  }

\begin{document}

\maketitle
\thispagestyle{firstheader}

\input{figures/fig1_in_paper}

\input{sections/1_introduction}

\input{sections/2_related_work}
\input{sections/3_contracts}

\input{figures/fig2_in_paper}

\input{sections/4_verification_training}
\input{sections/5_experimental_setup}
\input{sections/6_results}
\input{sections/7_qualitative}
\input{sections/8_discussion}
\input{sections/7_conclusion}

\FloatBarrier
\setlength{\bibsep}{4pt plus 1pt minus 1pt}
\bibliography{references}
\bibliographystyle{assets/plainnat}

\end{document}

%% file: math_commands.tex
\newcommand{\Cstar}{\ensuremath{C^{\star}}\xspace}
\newcommand{\Chat}{\ensuremath{\widehat C}\xspace}
\newcommand{\Acompile}{\ensuremath{\mathcal{A}}\xspace}
\newcommand{\Rformat}{\ensuremath{R_{\mathrm{format}}}\xspace}
\newcommand{\Rplan}{\ensuremath{R_{\mathrm{plan}}}\xspace}
\newcommand{\Rclear}{\ensuremath{R_{\mathrm{clear}}}\xspace}
\newcommand{\Rexec}{\ensuremath{R_{\mathrm{exec}}}\xspace}
\newcommand{\Rglobal}{\ensuremath{R_{\mathrm{global}}}\xspace}

\newcommand{\Rqual}{\ensuremath{R_{\mathrm{quality}}}\xspace}

\DeclareMathOperator*{\clip}{clip}

%% file: results_current/numbers.tex
\newcommand{\CrossInteraction}{1.75}

%% file: sections/0_abstract.tex
Complex image editing requires faithful task interpretation and accurate
execution, but a failed edit does not reveal which responsibility failed.
We present DuplexBridge, which uses a unified model's own understanding
capability to enhance prompts for its own generation capability, without an
external understanding module. A source-grounded bridge specifies targets,
final-state changes, and preservation constraints. Understanding is supervised
on fidelity to the original request without observing the generated image;
generation is supervised on execution of the condition it actually receives.
Reciprocal trust checks selectively admit global task feedback to each role
while retaining independently valid local supervision.
We introduce CE-Bench, a 1,200-task benchmark measuring joint satisfaction of
task-derived change and preservation requirements. Across three model
configurations, the full system improves complete compliance over direct
editors by 21.75--25.75 percentage points, with gains also observed on
KRIS-Bench, RISEBench, ImgEdit, and GEdit-Bench. From a common supervised Wan-Image
initialization, gated joint training reaches 77.08\% complete compliance,
compared with 70.92\% without gates and 75.00\% when generation-local
supervision instead checks the original task with both gates retained.
These results support distinct local responsibilities and selective shared
feedback for joint understanding--generation adaptation.

%% file: figures/fig1_in_paper.tex
\begin{figure}[!t]
  \centering
  \includegraphics[width=\linewidth]{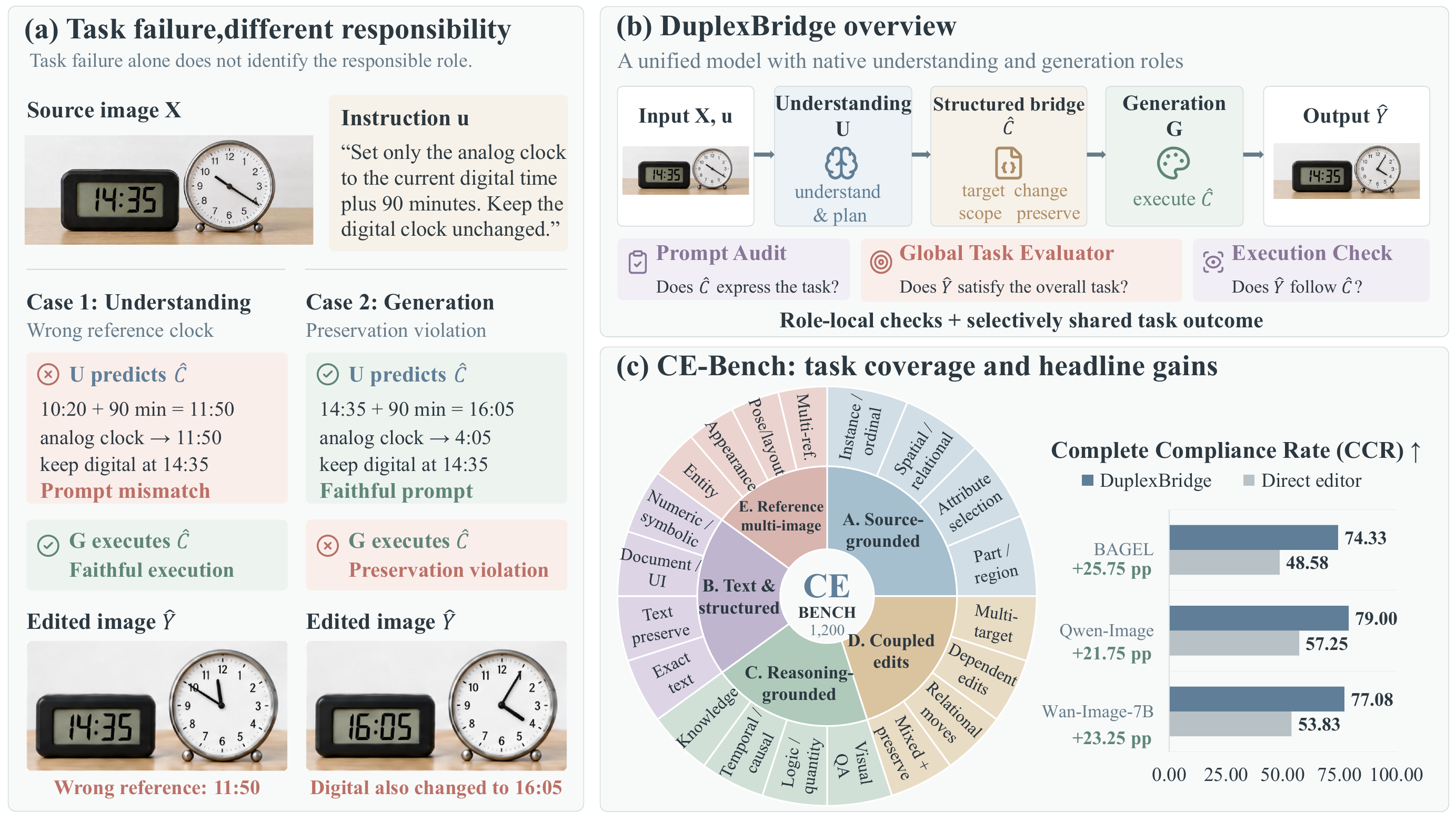}
  \caption{\textbf{DuplexBridge overview.} (a) Task failures with different local
responsibilities. (b) Native U--G communication and three distinct checks.
(c) CE-Bench coverage and full-system CCR gains, including PE, SFT, and RL.
Primary sectors encode 300/240/240/240/180 tasks; outer subdivisions illustrate
capabilities rather than subtype frequencies.}
  \label{fig:overview}
\end{figure}

%% file: sections/1_introduction.tex
\section{Introduction}
\label{sec:intro}

Complex image editing must bind changes to the correct source content, infer
final states, and preserve protected content. Setting only an analog clock to
the digital clock's time plus 90 minutes requires all three.

Image-grounded interpretation and joint optimization already support complex
editing \citep{fu2024mgie,huang2024smartedit}.
We study \emph{native prompt enhancement} (PE) in unified models
\citep{deng2025bagel,mao2026wanimage}: their own understanding capability, U,
constructs conditions for their own generation capability, G, without an
external understanding module. When adapting both roles, what should supervise
their local responsibilities, and when should they also learn from final task
outcomes?

A task-level score alone does not distinguish faithful task expression from
faithful execution. In \cref{fig:overview}(a), the digital reading 14:35 gives
an analog target of 4:05. U may instead use the wrong reference and specify
11:50, which G faithfully renders. Alternatively, U specifies 4:05 correctly,
but G changes the protected digital display. These failures motivate distinct
checks of prompt fidelity, received-condition execution, and task outcome.

Executable planning and two-role adaptation have precedents
\citep{zhou2026unifiedthinker,jiao2026thinkgen}.
DARS routes planner--renderer updates using task-derived checklists and
post-render planner diagnostics \citep{cao2026dars}; DDA-Thinker separately
evaluates plans and visual outcomes while keeping the editor frozen
\citep{ddathinker2026}. DuplexBridge combines distinct local supervision
references for both adapting roles with reciprocal checks that select only
their additional global-feedback losses.

The bridge specifies source-bound targets, changes, and preservation
(\cref{fig:overview}). G receives only the source and this condition, which
also defines G-local execution checks.

U-local checks compare the condition with the task without seeing the generated
image. A separate global reward
assesses the final edit against the task. Prompt trust gates this reward to G;
execution trust gates it to U. Valid local supervision remains when either
gate closes. Training alternates role updates; inference follows the native U-to-G path.

CE-Bench requires all task-derived change and preservation checks to pass;
these requirements are independent of the predicted bridge. Full-system
complete compliance rate (CCR) gains over direct editors span 21.75--25.75
points across three configurations, alongside public-benchmark improvements. From a common
supervised Wan-Image initialization, full reaches 77.08\% CCR versus 70.92\% without
gates and 75.00\% with original-task G-local and both gates retained.
Group-structure-matched random selection reaches 72.25\%, supporting semantic
selection in this setting.

Our contributions are:
\begin{enumerate}[leftmargin=*,nosep]
 \item A checkable bridge within a unified model that defines distinct local
 supervision targets for task expression and received-condition execution.
 \item A joint learning rule that retains valid local supervision while
 selectively sharing global task feedback through reciprocal partner-trust gates.
 \item CE-Bench, a 1,200-task evaluation of complete editing fulfillment,
 supported by public benchmarks and controlled comparisons of the feedback design.
\end{enumerate}

%% file: sections/2_related_work.tex
\section{Related Work}
\label{sec:related}
\label{sec:related-details}

\paragraph{Understanding--generation coordination.}
MGIE learns expressive instructions and jointly trains the multimodal language
model and image editor \citep{fu2024mgie}.
SmartEdit strengthens complex instruction editing with bidirectional interaction
between image features and MLLM outputs \citep{huang2024smartedit}.
These works establish understanding-guided editing and coupled adaptation;
SmartEdit's bidirectionality concerns feature interaction, whereas ours concerns
training feedback between the understanding and generation roles.
BAGEL unifies image understanding and generation; Uni-Edit studies editing-driven
adaptation \citep{deng2025bagel,zheng2026uniedit}.
Unified Thinker uses successive understanding and generation phases, ThinkGen
applies SepGRPO, and ThinkRL-Edit decouples the two roles' updates
\citep{zhou2026unifiedthinker,jiao2026thinkgen,thinkrledit2026}.
UniGRPO and BRAID further study joint reasoning and synthesis
\citep{liu2026unigrpo,hu2026braid}.
DuplexBridge focuses on the supervision references and feedback eligibility
within this co-adaptation: task fidelity, received-condition execution, and
task outcome train distinct loss components.
Here, \emph{native} understanding means that the unified model uses its own
understanding capability for prompt enhancement and passes the result to its
own generation capability, without an external understanding module in that
inference path.

\paragraph{Structured prompt enhancement and feedback.}
PromptEnhancer optimizes a rewriter with a frozen generator; its published
alignment reward evaluates generated images against the rewritten prompt
\citep{wang2026promptenhancer}. Thus, received-condition image assessment is
already an established component. Wan-Image and SenseNova-U1.5 use structured
editing conditions \citep{mao2026wanimage,diao2026sensenovau15}.
DDA-Thinker already assesses executable plans without observing generated
images and maintains separate cognitive and visual reward streams, while
keeping the editor frozen \citep{ddathinker2026}.
DARS addresses the same ambiguity between planning and execution failures.
It jointly trains a planner and renderer with an original-task checklist,
post-render planner diagnostics, prefix gating across plan fields, and
variance-based module weights \citep{cao2026dars}.
DuplexBridge combines different local references within co-adaptation:
U-local judges task fidelity without the rendered image, and G-local judges
execution against the condition actually received. Reciprocal partner trust
selects only the additional global losses, preserving valid local supervision
when global feedback is withheld. This combination makes local responsibility
and global-feedback eligibility explicit within the same training procedure.

\paragraph{Reward modeling and editing evaluation.}
SLD detects and corrects generation errors through an inference-time loop
\citep{wu2024sld}; DuplexBridge uses execution checking to determine which
feedback can train each role.
RewardDance uses comparative token probabilities; UnifiedReward-Think and
Z-Reward study reasoning-enhanced and distilled judgment
\citep{wu2025rewarddance,wang2025unifiedrewardthink,jin2026zreward}.
Our distilled RM supports the shared outcome signal through direct pairwise
scoring against fixed reference edits. TIFA and Davidsonian Scene Graph
motivate decomposed visual checking \citep{hu2023tifa,cho2024dsg}.
ImgEdit, GEdit, RISEBench, and KRIS-Bench assess complementary dimensions of
editing and reasoning \citep{imgedit2025,step1xedit2025,risebench2025,krisbench2025}.
CE-Bench emphasizes \emph{complete} compliance with source-grounded changes
and preservation constraints. Its requirements come from the original task,
independently of a model's predicted bridge, and evaluation accepts final images.

%% file: sections/3_contracts.tex
\section{A Structured Bridge for Understanding and Generation}
\label{sec:contracts}

\subsection{Native roles and the forward path}
Let $X$ contain the source image and any task-provided reference images, with
their roles identified, and let $u$ be the editing instruction. DuplexBridge
uses a unified model's own understanding role $U_\theta$ for prompt enhancement
(PE) and its own generation role $G_\phi$ for editing:
\begin{equation}
 (z_\theta,\Chat)\sim\pi_\theta(\cdot\mid X,u;\psi),\qquad
 \widehat Y\sim G_\phi(\cdot\mid X,\Chat;\psi).
 \label{eq:delta-contract}
\end{equation}
Here $z_\theta$ is internal interpretation in a \texttt{<think>} block and
$\Chat$ is the public structured bridge. Trainable role parameters
$\theta,\phi$ are disjoint; fixed pretrained parameters $\psi$ may include a
shared backbone. No external understanding module constructs $\Chat$.
Generation receives $X$ and $\Chat$ in a fresh context, without the original
instruction or U's internal reasoning.

\subsection{Making the editing requirements explicit}
\label{sec:private-construction}
\label{eq:student-planner}
U resolves target identities, reasoning-dependent answers, and terminal states
before constructing $\Chat$. The bridge binds changes to source entities,
separates editable from protected content, and specifies preservation
requirements and forbidden effects. Entailed requirements are made explicit
without additional edits; requested text retains its exact characters.
CE-v1 organizes these requirements as grouped JSON fields
\texttt{goal}, \texttt{scope}, \texttt{ops}, \texttt{preserve}, and \texttt{avoid}
(\cref{fig:bridge-uses}). The goal summarizes the operative fields without
adding obligations.

\label{sec:serialization}
\label{sec:executable-definition}
The public condition also defines the execution checks:
\begin{equation}
 D(\Chat)=(D_\Delta,D_{\rm inv}),\qquad
 Q_{\rm exec}=\Acompile(\Chat),
 \label{eq:final-contract}
\end{equation}
where $D_\Delta$ and $D_{\rm inv}$ are disjoint change and invariant groups.
A source resolver binds targets in $X$; fixed templates check applicable
initial states before output requirements. Repeated obligations are
deduplicated; overlapping preservation scopes are partitioned or deduplicated
after semantic validation. Template enumeration is deterministic; source
binding and visual judgment remain semantic operations.

The compiler uses the same public condition as G. Execution feedback evaluates
how faithfully G realizes that condition, while prompt feedback evaluates
its fidelity to the user's request.
U is initialized on accepted teacher responses containing internal interpretation
and public JSON before RL. \Cref{sec:schema,sec:initialization} describe the bridge semantics and the
initialization controls.

\subsection{Source binding, coverage, and validity}
\label{sec:schema}
\label{sec:compiler}
The bridge supports recoloring, replacement, insertion, removal, movement,
text transformation, and observable state changes. It expresses the requested
terminal state, including exact final text or a reasoning-derived value,
rather than an unresolved instruction to reason. Editable and protected source
content have disjoint roles. Targets must resolve to a source instance or a
determinate set; matching the edited output retains that source identity.

Each execution group conjoins source grounding, applicable initial-state
checks, and the corresponding output requirement. A failed or unresolved
prerequisite marks its group as non-passing once; training eligibility is
assessed separately. Preservation requirements reuse source evidence and are
deduplicated or partitioned when scopes overlap, so repeated wording does not
increase reward weight. These reduction rules also apply to public evaluation,
whose requirements come from the original task rather than the predicted bridge.

In the clock example, the source analog display reads 10:20 and the protected
digital display reads 14:35. The requested digital time plus 90 minutes yields
16:05, or 4:05 on the analog dial. The bridge specifies 4:05 as the analog
terminal state; reading the digital display supplies evidence without making
it an edit target. Its unchanged time and appearance, the analog dial and case,
and the surrounding scene remain preservation obligations.

The goal summary must be entailed by the operative requirements. Specification
validation checks operation support, target resolution, clause consistency,
and coverage of the goal by the operative requirements. Execution training
uses valid, resolvable specifications with confident judgments, including
confident execution failures. A confident prompt-fidelity judgment can
independently supervise U locally.
Private interpretation, expected judge answers, benchmark identifiers, and rewards
are excluded from the public condition passed to G.

%% file: figures/fig2_in_paper.tex
\begin{figure}[!htbp]
  \centering
  \includegraphics[width=\linewidth]{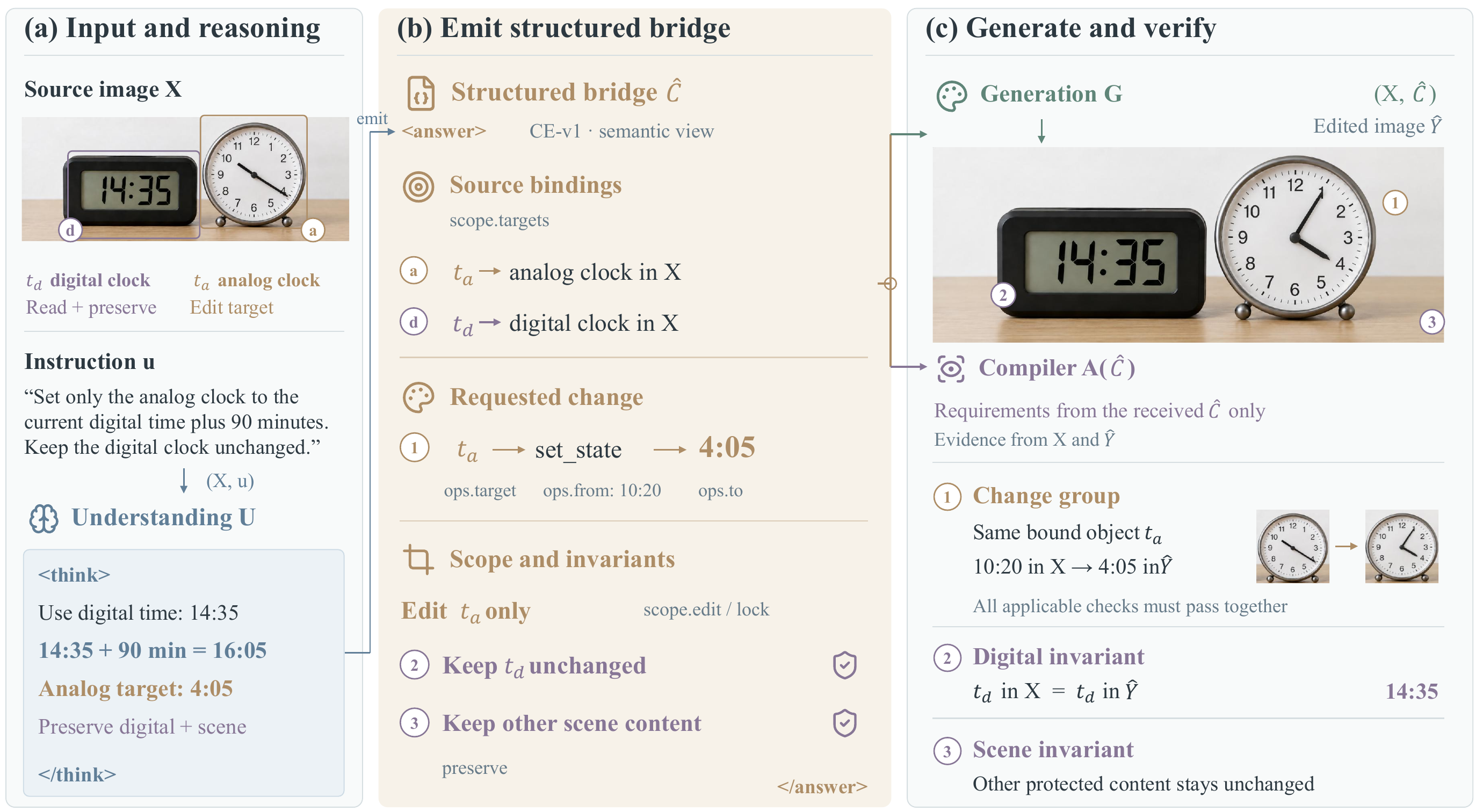}
  \caption{\textbf{From understanding to a checkable bridge.} U resolves the
clock task and separates the editable analog clock from protected digital and
scene content. Only the public condition enters G and the execution compiler;
change and invariant checks use that same condition.}
  \label{fig:bridge-uses}
\end{figure}

%% file: sections/4_verification_training.tex
\section{Coordinating Learning with Local and Global Feedback}
\label{sec:verification}
DuplexBridge separates three evaluations by the information they receive:
prompt fidelity uses $(X,u,\Chat)$, execution uses $(X,\Chat,\widehat Y)$,
and the global outcome uses $(X,u,\widehat Y)$
(\cref{fig:partner-feedback}). Local rewards train each role's responsibility;
partner trust determines when the joint outcome also supervises that role.

\subsection{Role-local rewards}
\begin{equation}
 R_U^{\rm loc}=F_U(X,u,\Chat),\qquad
 R_G^{\rm loc}=F_G(X,\Chat,\widehat Y).
 \label{eq:local-rewards}
\end{equation}
\paragraph{Understanding: prompt fidelity.}
A separately validated reference $\Cstar$ supports format, requirement-level
semantic precision/recall, and consistency/grounding judgments:
\begin{equation}
 R_U^{\rm loc}=s_C=(\Rformat+\Rplan+\Rclear)/3.
 \label{eq:audit-score}
\end{equation}
The plan score uses maximum one-to-one semantic matching between predicted and verified
requirements: target identity, operation, terminal value, and invariant scope
must agree. With $T$ matches among $P$ predicted and $R$ reference groups,
$\Rplan=2T/(P+R)$. Format is a binary schema-conformance check; clarity averages
a no-contradiction judgment and the resolved-source-binding fraction, with
malformed or empty predictions scored zero. This check never reads $\widehat Y$.
Unresolved judgments abstain. The verified reference is distinct from teacher
SFT targets (\cref{sec:private-construction}).

\paragraph{Generation: execution and local quality.}
For $Q_{\rm exec}=\{Q_1,\ldots,Q_K\}$, each change group combines target
grounding with its visual postconditions, while each distinct invariant forms
another group. With visual judge $V$ and check thresholds $\tau_q$,
\begin{align}
 \Rexec&=\frac1K\sum_{k=1}^K
 \prod_{q\in Q_k}\mathbf1[V(X,\widehat Y,q)\geq\tau_q],
 \label{eq:exec-score}\\
 R_G^{\rm loc}&=w_\Delta R_{{\rm exec},\Delta}
 +w_{\rm inv}R_{{\rm exec},{\rm inv}}+w_q\Rqual.
 \label{eq:generator-reward}
\end{align}
The nonnegative weights sum to one; change and invariant scores average their
respective groups, both of which must be nonempty for execution training.
Quality excludes instruction compliance, and preservation is counted once.
Invalid or unresolvable specifications and uncertain execution judgments are
excluded by the eligibility mask; low scores on valid, confidently judged
checks remain learning signals (\cref{sec:compiler}).

\subsection{Global task feedback}
\label{sec:independent-task}
A frozen pairwise reward model (RM) compares $\widehat Y$ with baseline edits
from a fixed, nonempty set $\mathcal R_{X,u}$ under the original instruction
and rubric $i$. It receives neither $\Chat$ nor U's reasoning.
Following the token-probability interface of RewardDance
\citep{wu2025rewarddance},
\begin{equation}
 \Rglobal=F_{\rm global}(X,u,\widehat Y;\mathcal R_{X,u})
 =\frac{1}{|\mathcal R_{X,u}|}\sum_{Y_r\in\mathcal R_{X,u}}
 P_\omega(\texttt{Yes}\mid X,u,\widehat Y,Y_r,i).
 \label{eq:global-score}
\end{equation}
Here \texttt{Yes} favors $\widehat Y$; its raw vocabulary probability gives a
relative preference score in $[0,1]$.
References are independent of candidate bridges and shared across each
rollout group. The direct-scoring RM distills a thinking teacher's judgment
distribution and remains frozen during U/G training; its formulation and
evaluation are in \cref{sec:reward-details,sec:reward-evaluation}.

\input{figures/fig3_in_paper}

\subsection{Bidirectional gates}
Prompt trust controls G-global feedback; execution trust controls U-global
feedback. Let $c_{\rm audit}$ indicate an available verified $\Cstar$ and a
sufficiently reliable prompt-fidelity judgment, and let $b_E,b_Q,b_F$ indicate
usable execution, quality, and global judgments. For a valid specification,
prompt trust requires
\begin{equation}
 \begin{aligned}
 v_C={}&\mathbf1[\Rformat=1]\mathbf1[\Rplan\geq\tau_p]
          \mathbf1[\Rclear\geq\tau_c]\\
       &\quad\times\mathbf1[\text{required targets resolved}].
 \end{aligned}
 \label{eq:prompt-trust}
\end{equation}
Here $\tau_p,\tau_c$ are the minimum plan and clarity scores; invalid
specifications have $b_E=v_C=0$. Unresolved source bindings or uncertain execution
judgments set $b_E=0$; unresolved required targets also give $v_C=0$.
Judgment usability is separate from success: $c_{\rm audit},b_E,b_Q$ require reliable judgments,
with reliability estimators and thresholds fixed using independent development
labels. The reciprocal gates and global-loss eligibility are
\begin{equation}
 \begin{aligned}
 m_{U\to G}&=c_{\rm audit}v_C,& e_G^{\rm glob}&=b_Fm_{U\to G},\\
 m_{G\to U}&=b_E\mathbf1[\Rexec\geq\tau_e],&
 e_U^{\rm glob}&=b_Fm_{G\to U}.
 \end{aligned}
 \label{eq:eligibility}
\end{equation}
Local eligibility is independent of these gates:
$e_U^{\rm loc}=c_{\rm audit}$ and $e_G^{\rm loc}=b_Eb_Q$.
Thus $b_E=0$ excludes G-local and U-global, while a reliable prompt judgment
can still train U locally. Confidently identified failures supply low local
rewards. Gate-only ablations bypass semantic trust while retaining
judgment-confidence and availability tests.

In \cref{fig:overview}(a), if G faithfully sets the analog clock to U's incorrect
11:50, G retains local supervision. Passed execution trust admits global feedback
to U; failed prompt trust excludes G-global. When U correctly specifies 4:05
but G changes the protected digital clock, passed prompt trust admits G-global;
U-global is excluded if execution trust fails.

\subsection{Alternating updates with four loss components}
\label{sec:posttraining}
\label{sec:optimization}
\label{sec:training}
Each role combines two separately normalized group-relative clipped losses:
\begin{equation}
 \begin{aligned}
 \mathcal L_U&=\mathcal L_{\rm GR}^{U,{\rm loc}}(R_U^{\rm loc};e_U^{\rm loc})
 +\lambda_U\mathcal L_{\rm GR}^{U,{\rm glob}}(\Rglobal;e_U^{\rm glob}),\\
 \mathcal L_G&=\mathcal L_{\rm GR}^{G,{\rm loc}}(R_G^{\rm loc};e_G^{\rm loc})
 +\lambda_G\mathcal L_{\rm GR}^{G,{\rm glob}}(\Rglobal;e_G^{\rm glob}).
 \end{aligned}
 \label{eq:constructor-reward}
\end{equation}
For U updates, freeze G, sample complete U responses, and generate several
images per response. Average global scores over eligible images within each
response, then weight eligible responses equally; a response with none can
still contribute to U-local. For G updates, freeze U and compare stochastic
image trajectories conditioned on one bridge.

For component $t\in\{U{\rm loc},U{\rm glob},G{\rm loc},G{\rm glob}\}$,
let $I_t=\{i:e_{ti}=1\}$ and compute its reward mean $\mu_t$ and standard
deviation $\sigma_t$ only on $I_t$. For an eligible, nondegenerate group,
\begin{equation}
 \begin{aligned}
 A_{ti}&=\frac{R_{ti}-\mu_t}{\sigma_t+\epsilon},\\
 \mathcal L_t&=\frac{1}{|I_t|}\sum_{i\in I_t}
 \left[-\min\{\rho_{ti}A_{ti},
 \clip(\rho_{ti},1-\varepsilon,1+\varepsilon)A_{ti}\}
 +\beta_t\mathcal D_{ti}\right].
 \end{aligned}
 \label{eq:masked-policy-loss}
\end{equation}
Here $\rho_{ti}$ is the current-to-behavior policy likelihood ratio and
$\mathcal D_{ti}$ the frozen-reference policy penalty. Each component normalizes
rewards and applies its KL penalty on its eligible set. Groups with
fewer than two eligible samples or zero reward variance are skipped for that
component, including KL. U uses text-policy likelihoods and G uses stochastic
flow trajectory likelihoods \citep{shao2024deepseekmath,flowgrpo2025}.
A sample excluded from global feedback can retain local supervision.

\subsection{Reasoning-distilled global judgment}
\label{sec:reward-details}
\label{sec:reward-distillation}
For pairwise input $x=(X,u,Y_A,Y_B,i)$, a frozen thinking teacher produces
reasoning $z$ and a judgment distribution $\mathbf q_T(x,z)$. The direct
student predicts $\mathbf p_\omega(x)$ from the same visual inputs, candidate
order, and rubric, without receiving or decoding the teacher's reasoning.
Both compare original-task fulfillment, preservation, and visual quality.
Predicted bridges, private U reasoning, human preference labels, and test
annotations are excluded from their inputs.

The RL score uses raw vocabulary probability $P_{\texttt{Yes}}$, whereas the
binary evaluation probability conditions on the two judgment tokens:
\begin{equation}
 r_{\rm raw}=P_{\texttt{Yes}},\qquad
 r_{\rm bin}=\frac{P_{\texttt{Yes}}}{P_{\texttt{Yes}}+P_{\texttt{No}}}.
 \label{eq:reward-probabilities}
\end{equation}
For raw-score distillation, the common outcome space retains the masses for
\texttt{Yes}, \texttt{No}, and all other vocabulary tokens. Soft judgment
distillation optimizes
\begin{equation}
 \mathcal L_{\rm KD}=\mathbb E_{x,z}
 D_{\rm KL}\!\left(\operatorname{sg}[\mathbf q_T(x,z)]
 \Vert\mathbf p_\omega(x)\right),
 \label{eq:reward-kd}
\end{equation}
with only the student updated using judgment-level supervision.
In expectation, minimizing this loss targets the frozen teacher's marginal
judgment distribution $\mathbb E_z[\mathbf q_T(x,z)]$.
After distillation, the RM stays frozen during U/G training and
supplies only global feedback. Independent preference labels, candidate-order
swaps, and difficult-pair evaluation assess its reliability
(\cref{sec:reward-experiments}).

%% file: figures/fig3_in_paper.tex
\begin{figure}[!htbp]
  \centering
  \includegraphics[width=\linewidth]{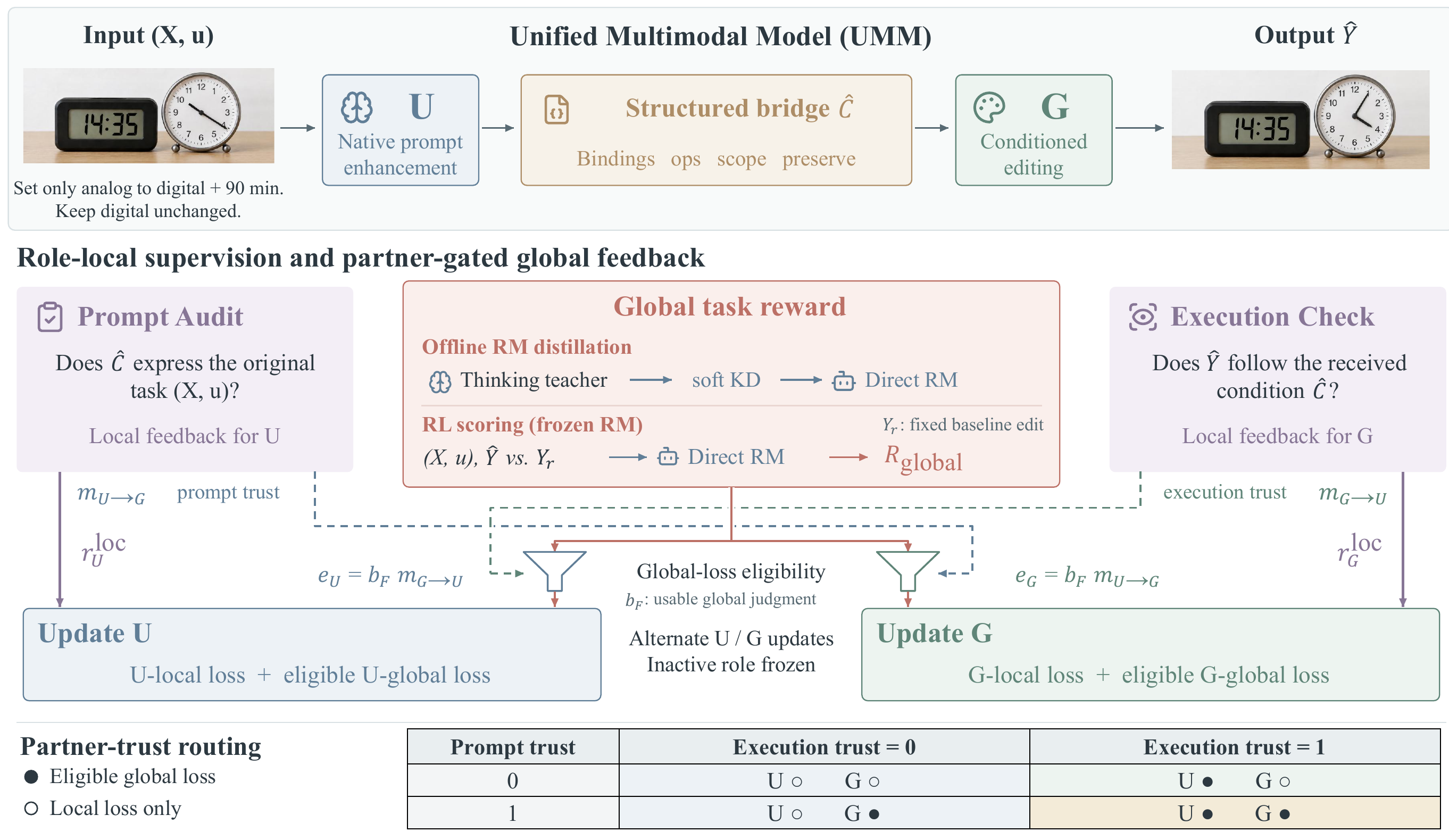}
  \caption{\textbf{Bidirectional partner-gated feedback.} Execution trust gates
U-global; prompt trust gates G-global. Local feedback retains independent
validity masks. The diagram abbreviates $e_U^{\rm glob},e_G^{\rm glob}$ as
$e_U,e_G$ and local rewards as $r_U^{\rm loc},r_G^{\rm loc}$.
The routing matrix assumes usable judgments. $Y_r$ is a baseline edit,
distinct from task-provided reference images.}
  \label{fig:partner-feedback}
\end{figure}

%% file: sections/5_experimental_setup.tex
\section{Experimental Setup}
\label{sec:setup}

\subsection{Models and baselines}
Image editing uses BAGEL, Qwen-Image-Edit-2511 (Qwen-2511), and Wan-Image-7B, with
mechanism ablations on Wan-Image. Each model supplies its own U and G for native PE.
Qwen-Image labels in the figures denote Qwen-2511.
Direct baselines edit the original instruction; full systems add native PE,
structured supervised initialization, and joint RL.
$B_0$ denotes the common supervised initialization before RL.
U-only, G-only, local-only, and joint variants are independently trained from
$B_0$, with common supervision and reference access.
Unless varied explicitly, global feedback uses the reasoning-distilled RM.
RM replacement experiments use equal update steps and rollout budgets.
The controls distinguish complete-system gains from supervised initialization,
communication, and RL effects (\cref{sec:initialization,sec:interfaces,sec:role-ablation}).

\subsection{CE-Bench and complete task fulfillment}
\label{sec:benchmark}
\label{sec:data}
Our 1,200-task benchmark tests whether target binding, terminal-state changes,
and preservation constraints are jointly satisfied. \Cref{sec:taxonomy} gives its five-category taxonomy.
Evaluation uses the original task, source, required reference images, and final
output; requirements are independent of the model's predicted bridge.
For item $n$, let $a_{nk}$ denote a binary pass for requirement group $k$,
and let $P_n$ be the mean pass rate over its preservation groups. We report
\begin{equation}
 \mathrm{AC}=\frac1N\sum_n\frac1{K_n}\sum_k a_{nk},\qquad
 \mathrm{CCR}=\frac1N\sum_n\prod_k a_{nk},\qquad
 \mathrm{Pres}=\frac1N\sum_n P_n.
 \label{eq:compliance}
\end{equation}
All metrics are percentages and weight items equally; failed outputs remain in
the denominator. CCR requires complete compliance with every change and
preservation group.
A change group includes target grounding and the required terminal state;
each distinct preservation obligation counts once. Correctly changing
the analog clock but altering the protected digital display fails CCR despite
passing the change check. These task-derived groups differ from the
predicted-condition groups used for G-local training.

Public obligations are derived
from the source, task-provided references, and original instruction before
predictions are assessed. Independently verified $C^\star$ supports prompt
checks; accepted teacher responses supply SFT targets.

\input{sections/benchmark_taxonomy}

\subsection{Public benchmark protocols}
\label{sec:public-protocols}
We report the KRIS-Bench composite, RISEBench-360 completion accuracy,
ImgEdit's 1--5 score, and GEdit-Bench English Full-set $G_O$ (0--10)
\citep{krisbench2025,risebench2025,imgedit2025,step1xedit2025,cao2026dars}.
Cross-system comparisons evaluate complete configurations; Wan-Image ablations test
components within one configuration.

RISEBench-360 has 85 temporal, 90 causal, 100 spatial, and 85 logical tasks.
Overall accuracy weights the four category accuracies by these task counts.

KRIS measures knowledge-dependent editing through a composite score. GEdit EN
follows the English Full-set $G_O$ column in DARS Table~S5. All benchmarks use
their native metric scales.

%% file: sections/benchmark_taxonomy.tex
\subsection{Taxonomy and task allocation}
\label{sec:taxonomy}
CE-Bench assigns each of its 1,200 tasks one primary category and a secondary
capability category (\cref{tab:ce-taxonomy}). CE-v1 in \cref{sec:schema} names
the prompt schema, not a benchmark version.

\begin{table}[!htbp]
\centering\footnotesize
\caption{\textbf{CE-Bench taxonomy.} The five primary categories are mutually
exclusive; the secondary categories describe capability coverage.}
\label{tab:ce-taxonomy}
\setlength{\tabcolsep}{5pt}
\renewcommand{\arraystretch}{1.12}
\begin{tabularx}{\linewidth}{@{}>{\raggedright\arraybackslash}p{.25\linewidth}r>{\raggedright\arraybackslash}X@{}}
\toprule
Primary category & $N$ & Secondary categories \\
\midrule
A. Source-Grounded & 300 &
A1 Instance / Ordinal Grounding; A2 Spatial / Relational Grounding;
A3 Attribute-Conditioned Selection; A4 Part / Region Grounding. \\
\addlinespace
B. Text \& Structured Content & 240 &
B1 Exact Text Editing; B2 Multi-Text \& Text Preservation;
B3 Document / UI Layout Editing; B4 Numeric / Symbolic / Structured Visual Editing. \\
\addlinespace
C. Reasoning-Grounded & 240 &
C1 Visual Question-to-Edit; C2 Logical / Spatial / Quantitative Reasoning;
C3 Temporal / Causal / Physical Inference; C4 Knowledge / Conceptual / Procedural Reasoning. \\
\addlinespace
D. Coupled \& Compositional & 240 &
D1 Multi-Target Composition; D2 Dependent / Coupled Edits;
D3 Relational Rearrangement; D4 Mixed Operations + Preservation. \\
\addlinespace
E. Reference-Conditioned Multi-Image & 180 &
E1 Entity / Subject Transfer; E2 Attribute / Appearance Transfer;
E3 Pose / Layout / Structural Transfer; E4 Multi-Reference Composition \& Reasoning. \\
\bottomrule
\end{tabularx}
\end{table}

\paragraph{Primary-label rules.}
Precedence is E$>$D$>$C$>$B$>$A. E requires an additional task-provided reference
image; otherwise D requires multiple principal changes or coupled editing
obligations, C requires inference to derive the target or terminal state,
B centers on exact text or structured content, and A on directly visible source
grounding. Thus, a supplied replacement number is B, while a number that must
be calculated is C; directly stated spatial relations are A, while relations
requiring inference are C. Ordinary preservation alone does not make a
single-change task D. Nonexclusive tags record grounding, reasoning, operations,
reference use, and preservation without counting tasks again in aggregate scores.

\paragraph{Source and reference roles.}
Each input $(X_{\rm src},\mathcal I_{\rm ref},u)$ requires one edited source
image. A--D have no additional reference images; E uses references for entity,
appearance, or structural information, with E4 requiring at least two.
The source defines the editable scene and protected content, while stable image
identifiers bind referenced attributes to their input images. Editors and
evaluators receive the same task images in the same order with these roles
preserved. These conditioning images are distinct from the baseline reference
\emph{edits} used as comparison outputs by the pairwise RM.

AC, CCR, and Pres. follow \cref{eq:compliance}. Overall scores weight category
means by their item counts (25\%, 20\%, 20\%, 20\%, 15\% for A--E), with failed
outputs retained in their assigned category and denominator.

%% file: sections/6_results.tex
\section{Experiments}
\label{sec:results}

\subsection{End-to-end editing performance}
\label{sec:results-public}
\label{sec:results-rl}

\input{results_current/ce_full}

DuplexBridge improves CE-Bench CCR by 25.75, 21.75, and 23.25 points on
BAGEL, Qwen-2511, and Wan-Image, respectively, alongside preservation
(\cref{tab:ce-full}).

\input{results_revision/public_benchmarks_main}

On public benchmarks (\cref{tab:public-benchmarks-main}), Wan-Image reaches
80.13/29.17 KRIS/RISE, gaining 19.29/16.11 points over its direct baseline.
Qwen-2511 reaches 82.16/31.11, versus DARS's 80.72/27.50.
Task-specific text-to-image retraining is evaluated on GenEval2, R2I, WISE-V,
and DPG in \cref{sec:t2i-results}.

\input{results_current/rise_breakdown}

\input{figures/cebench_qualitative_in_paper}

\paragraph{Qualitative behavior.}
The CE-Bench examples in \cref{fig:ce-qualitative} illustrate image-grounded
answers, spatial placement, and preservation. DuplexBridge inserts
``Jackrabbit'' and ``45'' while retaining the worksheet diagrams, places a
sliced piece on the upper-left plate, and combines a man, puppy, and
red-patterned toy rabbit with the woman's base scene. These examples accompany
the quantitative results in \cref{tab:ce-full}.
\Cref{sec:qualitative} adds reasoning and physical transformations on RISEBench
and KRIS-Bench, as well as additional CE-Bench edits.

\subsection{Supervised initialization controls}
\label{sec:initialization}
The main tables compare unadapted baselines with full DuplexBridge; the supervised
$B_0$ checkpoint uses native PE and structured SFT, before RL.
\Cref{tab:b0-ce} reports all three supervised checkpoints on CE-Bench and public benchmarks.
From $B_0$ to full DuplexBridge, CE CCR increases by 17.91, 16.92, and 17.66
percentage points for BAGEL, Qwen-2511, and Wan-Image-7B.
All three configurations improve the four public metrics and each RISE category
over their respective $B_0$ (\cref{tab:rise-components}).
For Qwen, preservation is 90.86 before adaptation, 90.14 at $B_0$, and 94.21
with DuplexBridge; the corresponding ImgEdit scores are 4.36, 4.28, and 4.43.

\input{results_current/b0_ce}

\subsection{Information-matched communication and verification}
\label{sec:interfaces}
\label{sec:check-construction}

\input{results_current/interface}

With task facts, training data, budget, and prediction source matched,
grouping raises CCR from 56.67 to 59.00 for key--value serialization and from
57.42 to 59.42 for JSON (\cref{tab:interface}). The smaller difference between
grouped syntaxes suggests that organization matters more than JSON syntax.
Removing an information-redundant goal or shuffling fields changes CCR little.
\Cref{tab:structure-reward} further evaluates structured communication during
gated RL.

\input{results_current/checks}

The construction comparison retains the same received contract, global
evaluator, and prompt checks (\cref{tab:verification}). Source fidelity measures
correct target, scope, state, and polarity relative to that contract;
original-task recall uses the task inventory, and duplication measures repeated
checks. These diagnostics cover all construction attempts, while human agreement
uses an adjudicated content-matched subset. Compiling checks improves source
fidelity and reduces duplication. The CCR column reports editing performance
after retraining with each construction method. \Cref{tab:reward-mechanism}
further compares the supervision reference used by G-local.

\subsection{Joint adaptation and supervision references}
\label{sec:role-ablation}

\input{results_current/training}

Joint ungated training reaches 70.92 CCR, above independently trained U-only
(66.67) and G-only (67.50) variants (\cref{tab:coadaptation}). Dual-role local-only
training retains both valid local losses and omits global losses and their KL
terms; adding selected global feedback raises CCR from 68.25 to 77.08.
All RL variants start from the common $B_0$ initialization.

UT-style staged and DARS-style routing transplant mechanisms into Wan-Image,
reaching 72.67 and 74.08 CCR. They complement the complete-system comparisons
in \cref{tab:public-benchmarks-main}. Full scores 4.36 on ImgEdit and 7.74 on
GEdit EN, compared with 4.37 and 7.71 for the DARS-style transplant.

\input{results_reward/mechanism_controls}

With both gates retained, G-local supervision referenced to the condition
actually received reaches 77.08 CCR, versus 75.00 when it instead checks the
original task (\cref{tab:reward-mechanism}). U-local and global feedback retain
the original task reference.

\subsection{Selecting reliable global feedback}
\label{sec:added-ablations}
\label{sec:gating}

\input{results_current/gate_endpoints}

Both prompt and execution trust contribute: either gate alone reaches
74.17 or 73.33 CCR, compared with 77.08 for both and 70.92 with neither
(\cref{tab:gates}). Gate ablations bypass semantic trust while preserving
judgment-confidence and availability tests and both local losses.
Soft semantic weighting reaches 76.33. Annotation-assisted oracle gates
use additional adjudicated annotations and reach 78.67 CCR, 81.07 KRIS,
and 30.00 RISE.

Semantic gating exceeds group-structure-matched random selection by 4.83
points (77.08 versus 72.25). Count-matched random selection reaches 71.33.
Group matching additionally accounts for eligible group sizes and the
availability of accepted images per prompt candidate. The gains over both
random controls support selecting feedback by semantic trust.

\input{results_added/mechanism}

Flat and structured conditions retain the same task facts, local checks,
initialization, data, and training budget; grouping is the varied factor.
With $(r_{00},r_{01},r_{10},r_{11})=(67.42,70.75,70.92,77.08)$ for
structure $s\in\{0,1\}$ and gating $g\in\{0,1\}$, the endpoint interaction is
\begin{equation}
 \Delta_{\rm SG}=(r_{11}-r_{10})-(r_{01}-r_{00})=6.16-3.33=2.83
 \quad\text{percentage points}.
 \label{eq:structure-gate-interaction}
\end{equation}
This positive interaction supports the complementary contributions of
structured communication and semantic gating.

\input{results_current/gate_diagnostics}

On the same 6,000 fixed rollouts, the wrong-prompt rate among eligible
G-global samples falls from 18.59\% to 2.90\%, and the unexecuted-prompt rate
among eligible U-global samples falls from 27.41\% to 4.60\%.
Labels combine external-VLM judgments and human annotation.
Coverage divides accepted counts by all 6,000 rollouts; error rates divide
by the respective accepted counts.

\subsection{Checkpoint cross-pairing}
\label{sec:partners}

\input{results_current/pairs}

U0/G0 are the shared B0 states; Ut/Gt are checkpoints from the same full run.
The CCR difference-in-differences is
\begin{equation}
 77.08-65.92-68.83+59.42=\CrossInteraction
 \quad\text{percentage points}.
 \label{eq:cross-interaction}
\end{equation}
The positive interaction indicates complementary adaptation of the two roles.
KRIS and RISE also improve under the learned pair, as reported in
\cref{tab:cross-pairing}.

\input{sections/requirement_analysis}

\subsection{Reward quality, cost, and downstream effects}
\label{sec:reward-evaluation}
\label{sec:reward-experiments}
We compare base direct scoring (M0), explicit thinking (M1), preference SFT
(M2), decision--reason training with direct deployment (M3), hard judgment
KD (M4), and soft judgment KD (M5). The main experiments use M5.

\paragraph{Judgment quality and reliability.}
\Cref{tab:reward-quality} reports HPA, difficult-pair accuracy, NLL, Brier score,
SwapErr, and A/B gap. The binary metric specification uses
$s=P_{\texttt{Yes}}/(P_{\texttt{Yes}}+P_{\texttt{No}})$:
\begin{equation}
 \begin{aligned}
 \mathrm{NLL}&=-\frac1N\sum_n[y_n\log s_n+(1-y_n)\log(1-s_n)],\\
 \mathrm{Brier}&=\frac1N\sum_n(s_n-y_n)^2,\qquad
 \mathrm{SwapErr}=\frac1N\sum_n|s(x_n)+s(x_n^{\rm swap})-1|.
 \end{aligned}
 \label{eq:reward-metrics}
\end{equation}
HPA measures agreement with human preference. The A/B gap is
$100|\mathrm{Acc}_A-\mathrm{Acc}_B|$ percentage points, comparing accuracy
with the preferred candidate in the two input positions. NLL, Brier, and
SwapErr characterize probability quality and sensitivity to candidate order.

\input{results_reward/quality}

\input{results_reward/hard_subsets}

Soft KD improves difficult-pair accuracy from 66.5\% to 74.8\%, approaching
explicit thinking's 75.8\% at $1.0\times$ rather than $6.0\times$ online latency
(\cref{tab:reward-quality}). Gains over base direct span all five difficult-pair
categories (\cref{tab:reward-hard}); M5 matches M1 on target/ordinal pairs and
exceeds it on preservation damage, while M1 leads on the other three categories.

\paragraph{External reward benchmarks.}
\Cref{tab:reward-external} keeps MMRB2 image-editing pair accuracy
\citep{hu2025mmrb2} separate from EditReward-Bench candidate-count-dependent
ranking scores \citep{editreward2025}. M5 improves MMRB2 from 66.2 to 70.6,
and EditReward $K=2,3,4$ from 57.2/35.9/10.8 to 61.0/43.5/15.8.
M1 achieves the highest scores on these benchmarks, while M5 provides direct
scoring at lower online cost.

\input{results_reward/external}

\paragraph{Downstream reward--gate interaction.}
\Cref{tab:reward-downstream} reports all five configurations. D0--D3 cross
base/distilled direct reward with ungated/gated learning; D4 uses explicit
thinking with gates. Relative to base direct, M5 raises CCR by 1.84 points
without gates and 2.83 with gates. Gating adds 5.17 points under M0 and 6.16
under M5, yielding a reward--gate interaction of $6.16-5.17=0.99$ points.
M1 with gates reaches 77.50 CCR,
0.42 above M5, while M5 has slightly higher preservation (93.86 versus 93.73).
D0--D4 use equal update steps and rollout budgets.

\input{results_reward/downstream}

\input{sections/human_evaluation}

\subsection{Text-to-image adaptation}
\label{sec:t2i-results}

\input{results_t2i/generalization}

DuplexBridge-T2I retrains both roles: U specifies requirements from text and G
synthesizes from that condition. The Qwen T2I backbone is Qwen-Image;
BAGEL's baseline uses native CoT. All three configurations improve on all five
metrics (\cref{tab:t2i-generalization}).

GenEval2 Soft-TIFA averages per-prompt arithmetic (AM) or geometric (GM)
means of soft VQA scores \citep{kamath2025geneval2}. R2I tests reasoning
\citep{chen2025r2i}, WISE-V world-knowledge consistency \citep{niu2026wise},
and DPG dense prompt following \citep{hu2024ella}. R2I and WISE-V use the
displayed $\times100$ scales.

\FloatBarrier

%% file: results_current/ce_full.tex
\begin{table}[!htbp]
\centering\footnotesize\setlength{\tabcolsep}{3pt}
\caption{\textbf{CE-Bench comparison.} A--E are category CCR (\cref{tab:ce-taxonomy}); scores are percentages. Full systems include native PE, structured SFT, and gated RL. Dark/light pink marks the best/second-best values within each model group, including ties.}
\label{tab:ce-full}
\begin{tabularx}{\linewidth}{@{}Xrrrrrrrr@{}}
\toprule
Model & A & B & C & D & E & AC $\uparrow$ & Pres. $\uparrow$ & CCR $\uparrow$ \\
\midrule
\rowcolor{tableGroup}
\multicolumn{9}{l}{\textbf{Commercial Models}} \\
GPT-Image-1.5 & \BestScore{83.00} & \BestScore{78.33} & \SecondScore{72.92} & \SecondScore{82.50} & \SecondScore{68.89} & \SecondScore{93.47} & \SecondScore{94.12} & \SecondScore{77.83} \\
Gemini-3-pro-image-preview & \SecondScore{82.33} & 77.08 & \BestScore{81.67} & \BestScore{84.17} & \BestScore{72.78} & \BestScore{94.08} & 93.87 & \BestScore{80.08} \\
Seedream 4.5 & 80.33 & \SecondScore{77.50} & 62.92 & 72.50 & 58.89 & 91.36 & \BestScore{94.43} & 71.50 \\
Gemini-2.5-Flash-Image & 73.00 & 69.58 & 64.17 & 70.83 & 59.44 & 89.67 & 90.72 & 68.08 \\
\midrule
\rowcolor{tableGroup}
\multicolumn{9}{l}{\textbf{Generation-only Editors}} \\
FLUX.1-Kontext-dev (direct) & 58.33 & 43.33 & 30.42 & 43.75 & 33.33 & 75.62 & 86.31 & 43.08 \\
FLUX.2-klein-9B (direct) & \BestScore{75.67} & \SecondScore{64.58} & \SecondScore{47.08} & \SecondScore{60.00} & \BestScore{50.56} & \BestScore{85.92} & \SecondScore{92.87} & \BestScore{60.83} \\
LongCat-Image-Edit (direct) & \SecondScore{74.33} & \BestScore{65.42} & \BestScore{48.33} & \BestScore{60.42} & \SecondScore{47.22} & \SecondScore{85.74} & \BestScore{93.16} & \SecondScore{60.50} \\
\midrule
\rowcolor{tableGroup}
\multicolumn{9}{l}{\textbf{Planner + Generator}} \\
RePlan + Qwen-2511 & 70.33 & 58.75 & 52.92 & 59.58 & 49.44 & 84.62 & 90.41 & 59.25 \\
PromptEnhancerV2 + Qwen-2511 & 76.33 & 68.75 & 60.83 & 66.67 & 56.11 & 88.62 & 91.26 & 66.75 \\
ThinkRL-Edit (DARS config.) & 76.00 & 64.58 & 65.42 & 70.42 & 57.78 & 89.06 & 91.84 & 67.75 \\
EditThinker + Qwen-2511 & \SecondScore{77.33} & 67.08 & 63.33 & 70.00 & 60.56 & 89.48 & 92.18 & 68.50 \\
DDA-Thinker-32B + Qwen-2511 & \BestScore{79.00} & \SecondScore{71.25} & \SecondScore{70.83} & \SecondScore{75.42} & \SecondScore{64.44} & \SecondScore{91.17} & \SecondScore{93.10} & \SecondScore{72.92} \\
DARS & \BestScore{79.00} & \BestScore{72.50} & \BestScore{71.67} & \BestScore{76.25} & \BestScore{66.11} & \BestScore{91.36} & \BestScore{93.18} & \BestScore{73.75} \\
\midrule
\rowcolor{tableGroup}
\multicolumn{9}{l}{\textbf{Unified MLLMs}} \\
UniREdit-BAGEL & 72.00 & 57.08 & 55.00 & 62.92 & 47.78 & 85.68 & 89.34 & 60.17 \\
UniReason 1.0 & 74.00 & 62.50 & 60.42 & 67.92 & 53.33 & 87.81 & 90.26 & 64.67 \\
BAGEL & 64.00 & 43.75 & 40.00 & 51.67 & 36.67 & 80.80 & 86.72 & 48.58 \\
BAGEL + DuplexBridge & 79.00 & 70.83 & 74.17 & \SecondScore{77.08} & \SecondScore{67.78} & 91.27 & 92.74 & 74.33 \\
Qwen-2511 & 71.67 & 56.25 & 51.25 & 59.17 & 40.00 & 84.30 & 90.86 & 57.25 \\
Qwen-2511 + DuplexBridge & \SecondScore{82.00} & \SecondScore{78.33} & \BestScore{78.33} & \BestScore{80.83} & \BestScore{73.33} & \BestScore{93.16} & \BestScore{94.21} & \BestScore{79.00} \\
Wan-Image-7B & 68.33 & 52.08 & 46.67 & 56.67 & 37.78 & 82.70 & 88.47 & 53.83 \\
Wan-Image-7B + DuplexBridge & \BestScore{85.67} & \BestScore{79.17} & \SecondScore{77.08} & 67.08 & \BestScore{73.33} & \SecondScore{92.31} & \SecondScore{93.86} & \SecondScore{77.08} \\
\bottomrule
\end{tabularx}
\end{table}

%% file: results_revision/public_benchmarks_main.tex
\begin{table}[!htbp]
\centering\small\setlength{\tabcolsep}{4pt}
\caption{\textbf{Public editing benchmarks.} Higher is better; metric scales
follow \cref{sec:setup}. Dark/light pink marks the best/second-best value per
column among displayed systems.}
\label{tab:public-benchmarks-main}
\begin{tabular}{@{}lrrrr@{}}
\toprule
\rowcolor{tableGroup}
Method & KRIS $\uparrow$ & RISE $\uparrow$ & ImgEdit $\uparrow$ & GEdit EN $\uparrow$ \\
\midrule
PromptEnhancerV2 & 71.54 & 18.90 & 4.18 & 7.59 \\
DARS & \SecondScore{80.72} & 27.50 & \SecondScore{4.39} & \SecondScore{7.86} \\
\midrule
BAGEL & 56.21 & 5.80 & 3.45 & 6.40 \\
\quad + DuplexBridge & 76.48 & 26.11 & 4.04 & 7.41 \\
Qwen-2511 & 64.42 & 17.50 & 4.36 & 6.97 \\
\quad + DuplexBridge & \BestScore{82.16} & \BestScore{31.11} & \BestScore{4.43} & \BestScore{7.89} \\
Wan-Image-7B & 60.84 & 13.06 & 4.09 & 6.88 \\
\quad + DuplexBridge & 80.13 & \SecondScore{29.17} & 4.36 & 7.74 \\
\bottomrule
\end{tabular}
\end{table}

%% file: results_current/rise_breakdown.tex
\begin{table}[!htbp]
\centering\small\setlength{\tabcolsep}{5pt}
\caption{Native UMM configurations on RISEBench-360. Qwen-2511 denotes Qwen-Image-Edit-2511. Overall is task-weighted across 85 temporal, 90 causal, 100 spatial, and 85 logical items. Dark/light pink mark the best/second-best value in each column, including ties.}
\label{tab:rise-components}
\begin{tabular}{@{}lrrrrr@{}}
\toprule
Method & Temporal & Causal & Spatial & Logical & Overall \\
\midrule
BAGEL (B0) & 14.12 & 23.33 & 23.00 & 5.88 & 16.94 \\
BAGEL + DuplexBridge & 24.71 & 33.33 & \SecondScore{31.00} & 14.12 & 26.11 \\
Qwen-2511 (B0) & 21.18 & 26.67 & 29.00 & 9.41 & 21.94 \\
Qwen-2511 + DuplexBridge & \BestScore{31.76} & \BestScore{38.89} & \BestScore{34.00} & \BestScore{18.82} & \BestScore{31.11} \\
Wan-Image-7B (B0) & 16.47 & 23.33 & 24.00 & 7.06 & 18.06 \\
Wan-Image-7B + DuplexBridge & \SecondScore{29.41} & \SecondScore{35.56} & \BestScore{34.00} & \SecondScore{16.47} & \SecondScore{29.17} \\
\bottomrule
\end{tabular}
\end{table}

%% file: figures/cebench_qualitative_in_paper.tex
\begin{figure}[!htbp]
\centering
\includegraphics[page=3,trim=100bp 40bp 32bp 15bp,clip,width=\linewidth]{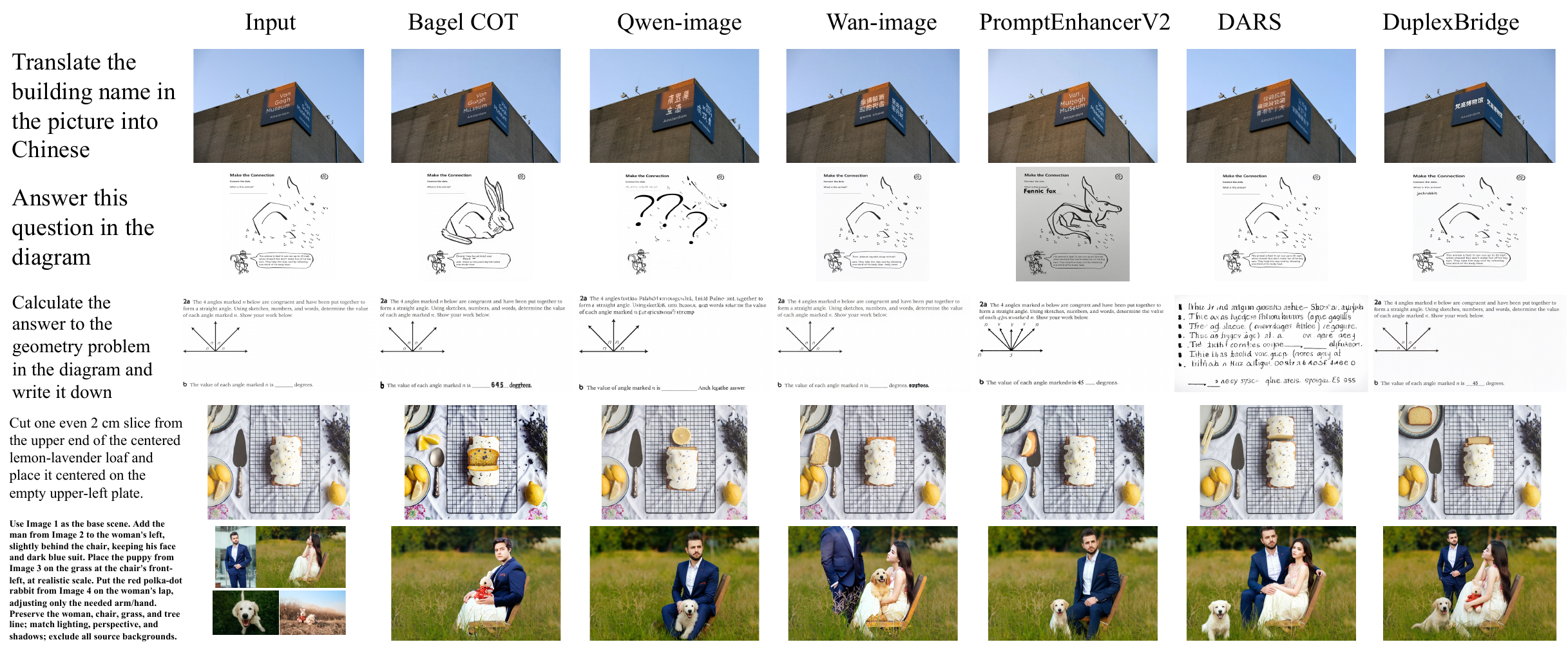}
\caption{\textbf{Qualitative comparison on CE-Bench.} Rows cover
sign translation, image-grounded question answering, geometric reasoning,
spatially specified slicing, and composition from multiple references.
The geometry example requires an answer within the existing diagram; the
final row combines entity placement with preservation of the base scene.}
\label{fig:ce-qualitative}
\end{figure}

%% file: results_current/b0_ce.tex
\begin{table}[!htbp]
\centering\footnotesize\setlength{\tabcolsep}{5pt}
\caption{\textbf{Supervised initialization controls.} $B_0$ uses native PE and structured SFT, before RL. A--E follow \cref{tab:ce-taxonomy}; public metrics follow \cref{sec:public-protocols}. Dark/light pink mark the best/second-best values, including ties.}
\label{tab:b0-ce}
\textbf{CE-Bench}\par\smallskip
\begin{tabular}{@{}lrrrrrrrr@{}}
\toprule
Model & A & B & C & D & E & AC & Pres. & CCR \\
\midrule
BAGEL (B0) & 68.67 & 49.58 & 47.92 & \SecondScore{62.92} & 47.78 & 83.91 & 88.63 & 56.42 \\
Qwen-2511 (B0) & \BestScore{73.00} & \BestScore{61.25} & \BestScore{55.00} & \BestScore{64.58} & \BestScore{51.11} & \BestScore{86.62} & \BestScore{90.14} & \BestScore{62.08} \\
Wan-Image-7B (B0) & \SecondScore{70.00} & \SecondScore{57.92} & \SecondScore{52.50} & 62.08 & \SecondScore{49.44} & \SecondScore{85.28} & \SecondScore{89.21} & \SecondScore{59.42} \\
\bottomrule
\end{tabular}
\par\medskip
\textbf{Public benchmarks}\par\smallskip
\begin{tabular}{@{}lrrrr@{}}
\toprule
Model & KRIS & RISE & ImgEdit & GEdit EN \\
\midrule
BAGEL (B0) & 64.23 & 16.94 & 3.74 & 6.93 \\
Qwen-2511 (B0) & \BestScore{70.36} & \BestScore{21.94} & \BestScore{4.28} & \BestScore{7.28} \\
Wan-Image-7B (B0) & \SecondScore{66.82} & \SecondScore{18.06} & \SecondScore{4.16} & \SecondScore{7.09} \\
\bottomrule
\end{tabular}
\end{table}

%% file: results_current/interface.tex
\begin{table}[!htbp]
\centering\small\setlength{\tabcolsep}{5pt}
\caption{Wan-Image interface ablations at SFT. Rows compare prose, flat and grouped structures, and CE-v1 variants without the redundant goal or with shuffled fields. AC and CCR are percentages.}
\label{tab:interface}
\begin{tabular}{@{}lrr@{}}
\toprule
Interface & CE AC & CE CCR \\
\midrule
Fact-matched prose & 84.38 & 57.00 \\
Flat key-value & 84.16 & 56.67 \\
Flat JSON & 84.55 & 57.42 \\
Grouped key-value & 85.09 & 59.00 \\
CE-v1 grouped JSON & 85.28 & 59.42 \\
CE-v1 without redundant goal & 85.17 & 59.17 \\
CE-v1 shuffled fields & 85.02 & 58.92 \\
\bottomrule
\end{tabular}
\end{table}

%% file: results_current/checks.tex
\begin{table}[!htbp]
\centering\footnotesize\setlength{\tabcolsep}{5pt}
\caption{Wan-Image execution-check construction. Fidelity, task recall, and duplicates use all rubric attempts. Human agreement uses an adjudicated content-matched subset. The final column reports CCR after retraining with each construction method.}
\label{tab:verification}
\begin{tabular}{@{}lrrrrr@{}}
\toprule
Construction & Fidelity & Recall & Dup. & Human agr. & Retrained CCR \\
\midrule
Same-contract prose rubric & 95.11 & 88.73 & 4.28 & 87.14 & 74.92 \\
Same-contract atomic reconstruction & 95.46 & 89.12 & 3.17 & 87.36 & 75.58 \\
Compiled execution checks & 98.21 & 91.68 & 0.73 & 87.92 & 77.08 \\
\bottomrule
\end{tabular}
\end{table}

%% file: results_current/training.tex
\begin{table}[!htbp]
\centering\footnotesize\setlength{\tabcolsep}{3.5pt}
\caption{\textbf{Wan-Image-7B training-direction ablation.} All RL variants
start at $B_0$ with common supervision and reference access. Native PE omits
CE-v1 SFT; U-only and G-only are trained independently. UT/DARS-style rows
transplant mechanisms into Wan-Image. Dashes in the local-only row denote
unreported metrics. GEdit EN uses Full-set $G_O$.}
\label{tab:coadaptation}
\begin{tabular}{@{}lcrrrrrrr@{}}
\toprule
Variant & U/G RL & \makecell{CE\\AC} & \makecell{CE\\CCR} & Pres. & KRIS & RISE & ImgEdit & \makecell{GEdit\\EN} \\
\midrule
Direct, no PE & --/-- & 82.70 & 53.83 & 88.47 & 60.84 & 13.06 & 4.09 & 6.88 \\
Native PE & --/-- & 83.86 & 56.67 & 88.92 & 63.41 & 15.83 & 4.11 & 6.97 \\
$B_0$: matched SFT & --/-- & 85.28 & 59.42 & 89.21 & 66.82 & 18.06 & 4.16 & 7.09 \\
U-only RL & $\checkmark$/-- & 87.96 & 66.67 & 88.74 & 73.28 & 24.44 & 4.17 & 7.21 \\
G-only RL & --/$\checkmark$ & 88.61 & 67.50 & 92.28 & 71.94 & 21.94 & 4.31 & 7.52 \\
Dual-role local-only & $\checkmark$/$\checkmark$ & -- & 68.25 & -- & -- & -- & -- & -- \\
Joint RL, ungated & $\checkmark$/$\checkmark$ & 89.72 & 70.92 & 91.82 & 76.27 & 26.11 & 4.34 & 7.58 \\
UT-style staged & $\checkmark$/$\checkmark$ & 90.38 & 72.67 & 92.14 & 77.39 & 26.94 & 4.32 & 7.60 \\
DARS-style routing & $\checkmark$/$\checkmark$ & 90.94 & 74.08 & 92.57 & 78.68 & 27.78 & 4.37 & 7.71 \\
DuplexBridge, full & $\checkmark$/$\checkmark$ & 92.31 & 77.08 & 93.86 & 80.13 & 29.17 & 4.36 & 7.74 \\
\bottomrule
\end{tabular}
\end{table}

%% file: results_reward/mechanism_controls.tex
\begin{table}[!htbp]
\centering\footnotesize
\caption{\textbf{Additional mechanism endpoints on Wan-Image.} CCR is a percentage;
the final column gives full minus control (percentage points). Count-matched
and group-structure-matched random gates are separate controls. The G-local
comparison changes the reference for local supervision from the original task
to the received condition. These endpoints compare training variants.}
\label{tab:reward-mechanism}
\setlength{\tabcolsep}{4pt}
\begin{tabular}{@{}lrr@{}}
\toprule
Condition & CCR $\uparrow$ & Full minus control \\
\midrule
Count-matched random gate & 71.33 & 5.75 \\
Group-structure-matched random gate & 72.25 & 4.83 \\
Original-task G-local source & 75.00 & 2.08 \\
Full: semantic gates + received-condition G-local & 77.08 & --- \\
\bottomrule
\end{tabular}
\end{table}

%% file: results_current/gate_endpoints.tex
\begin{table}[!htbp]
\centering\small\setlength{\tabcolsep}{5pt}
\caption{Wan-Image global-reward gate ablations. The first four rows vary the two semantic trust decisions, with local supervision and reward-validity conditions retained. Oracle gates use extra annotations.}
\label{tab:gates}
\begin{tabular}{@{}lrrr@{}}
\toprule
Gate & CE CCR & KRIS & RISE \\
\midrule
Neither global gate & 70.92 & 76.27 & 26.11 \\
Prompt trust only & 74.17 & 78.62 & 28.06 \\
Execution trust only & 73.33 & 78.14 & 27.50 \\
Both global gates & 77.08 & 80.13 & 29.17 \\
Count-matched random & 71.33 & 76.48 & 26.11 \\
Soft semantic weighting & 76.33 & 79.76 & 28.89 \\
Annotation-assisted oracle gates & 78.67 & 81.07 & 30.00 \\
\bottomrule
\end{tabular}
\end{table}

%% file: results_added/mechanism.tex
\begin{table}[!htbp]
\centering\small
\caption{\textbf{Structure $\times$ global gating on Wan-Image.} CE CCR (\%)
with matched task facts, local checks, initialization, data, and training budget.}
\label{tab:structure-reward}
\begin{tabular}{@{}lrr@{}}
\toprule
Interface & Ungated & Gated \\
\midrule
Flat & 67.42 & 70.75 \\
Structured & 70.92 & 77.08 \\
\midrule
\multicolumn{3}{l}{Interaction: $+2.83$ points} \\
\bottomrule
\end{tabular}
\end{table}

%% file: results_current/gate_diagnostics.tex
\begin{table}[!htbp]
\centering\footnotesize\setlength{\tabcolsep}{5pt}
\caption{Global-feedback selection on a frozen pool of 6,000 rollouts. Coverage uses all rollouts; error rates use the corresponding global-eligible subset. Wrong-prompt and unexecuted-prompt labels combine external-VLM judgments and human annotation.}
\label{tab:gate-diagnostics}
\begin{tabular}{@{}lrrrr@{}}
\toprule
Gate & G-global cov. & Wrong / G-global & U-global cov. & Unexec. / U-global \\
\midrule
Neither semantic gate & 93.60 & 18.59 & 91.40 & 27.41 \\
Prompt trust only & 71.20 & 2.90 & 91.40 & 27.41 \\
Execution trust only & 93.60 & 18.59 & 64.80 & 4.60 \\
Both trust gates & 71.20 & 2.90 & 64.80 & 4.60 \\
Count-matched random & 71.20 & 18.12 & 64.80 & 27.11 \\
\bottomrule
\end{tabular}
\end{table}

%% file: results_current/pairs.tex
\begin{table}[!htbp]
\centering\small\setlength{\tabcolsep}{5pt}
\caption{Wan-Image checkpoint cross-pairing. U0/G0 are the common B0 initialization; Ut/Gt are the adapted checkpoints from the same full training run.}
\label{tab:cross-pairing}
\begin{tabular}{@{}lrrr@{}}
\toprule
Pair & CE CCR & KRIS & RISE \\
\midrule
U0 G0 & 59.42 & 66.82 & 18.06 \\
Ut G0 & 65.92 & 72.48 & 22.22 \\
U0 Gt & 68.83 & 73.94 & 24.17 \\
Ut Gt & 77.08 & 80.13 & 29.17 \\
\bottomrule
\end{tabular}
\end{table}

%% file: sections/requirement_analysis.tex
\subsection{Requirement load and joint fulfillment}
\label{sec:requirement-load}
We compare ungated and full Wan-Image outputs on all 1,200 CE-Bench tasks,
paired by task. Public load $K$ counts change and preservation groups from the
original task, independently of the predicted bridge; it differs from the
number of requested edits. Let $C$ and $P$ indicate that all change or all
preservation groups pass, respectively. Complete success is $CP$, while Pres.
in \cref{eq:compliance} averages preservation-group pass rates within each task.

\input{figures/requirement_load_joint_in_paper}

\input{results_analysis/requirement_load}

Full has higher CCR point estimates at every observed load, with an overall
paired gain of 6.17 points [4.50, 7.83] (\cref{tab:requirement-load}). At $K=9$,
the CCR gain is 3.33 points [$-0.56$, 7.22]; change all-pass falls from 73.33\% to
60.56\% ($-12.78$ points [$-17.78$, $-7.78$]), while preservation all-pass
rises from 49.44\% to 52.78\%.

Across all tasks, change all-pass rises from 80.42\% to 83.50\%, preservation
all-pass from 76.75\% to 81.25\%, and mean failed groups fall from 0.659 to
0.508. The joint states appear in \cref{fig:requirement-load-joint}.
Task-paired transitions include 92 joint failures becoming successes and
18 successes becoming failures; 73 tasks gain preservation all-pass and 19
lose it. At $K=7$, preservation-only failures increase from 16.11\% to 17.50\%;
at $K=9$, both-component failures increase from 26.67\% to 39.44\%.

Load strata also differ in category composition: $K=3$ has no D tasks, while
$K=9$ has no A tasks and includes 95/180 D tasks. Category C at $K=9$ has a
$-6.67$-point CCR contrast on 30 tasks.

Intervals use 50,000 paired task bootstrap resamples, preserving category
counts overall and within each reported load. The 2.5th and 97.5th percentiles
give pointwise 95\% intervals; contrasts are computed before rounding.

%% file: figures/requirement_load_joint_in_paper.tex
\begin{figure}[!htbp]
\centering
\includegraphics[width=\linewidth]{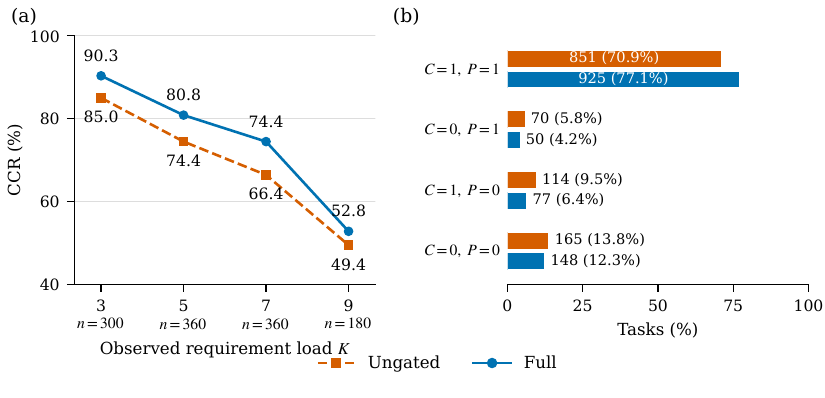}
\caption{\textbf{Requirement load and joint change--preservation outcomes.}
(a) CCR for full and ungated on the same 1,200 public tasks, grouped by their
observed total number $K$ of change and preservation requirements; each
stratum's task count is shown below its tick.
(b) Counts and percentages of the same tasks in the four joint states
(ungated above, full below), derived from task-level change and invariant
failure counts. $C=1$ means
every requested change passes, and $P=1$ means every preservation requirement
passes; $P$ is an all-pass indicator, distinct from the mean preservation
fraction reported as Pres. The fraction of tasks in the $C=P=1$ state equals
CCR.}
\label{fig:requirement-load-joint}
\end{figure}

%% file: results_analysis/requirement_load.tex
\begin{table}[!htbp]
\centering\footnotesize\setlength{\tabcolsep}{5pt}
\caption{\textbf{Complete and average compliance by public requirement load.} U/F denote ungated/full Wan-Image; CCR and AC are percentages. $\Delta$ is full minus ungated CCR in points, calculated before rounding. Pointwise paired bootstrap intervals follow the protocol in \cref{sec:requirement-load}.}
\label{tab:requirement-load}
\begin{tabular}{@{}lrrrrrrr@{}}
\toprule
$K$ & $N$ & U CCR & F CCR & $\Delta$ & 95\% interval & U AC & F AC \\
\midrule
3 & 300 & 85.00 & 90.33 & +5.33 & [+2.33, +8.33] & 95.00 & 96.78 \\
5 & 360 & 74.44 & 80.83 & +6.39 & [+3.33, +9.44] & 89.78 & 92.33 \\
7 & 360 & 66.39 & 74.44 & +8.06 & [+4.72, +11.39] & 88.53 & 92.58 \\
9 & 180 & 49.44 & 52.78 & +3.33 & [-0.56, +7.22] & 83.15 & 84.26 \\
All & 1200 & 70.92 & 77.08 & +6.17 & [+4.50, +7.83] & 89.72 & 92.31 \\
\bottomrule
\end{tabular}
\end{table}

%% file: results_reward/quality.tex
\begin{table}[!htbp]
\centering\footnotesize
\caption{\textbf{Reward-model judgment quality and cost.} HPA/Hard are accuracies (\%); A/B gap is in percentage points. NLL/Brier/SwapErr follow \cref{eq:reward-metrics}. Latency is one end-to-end pairwise call relative to M0; only M1 reasons online.}
\label{tab:reward-quality}
\setlength{\tabcolsep}{3.5pt}
\begin{tabular}{@{}lrrrrrrr@{}}
\toprule
RM & HPA $\uparrow$ & Hard $\uparrow$ & NLL $\downarrow$ & Brier $\downarrow$ & SwapErr $\downarrow$ & \makecell{A/B gap\\$\downarrow$} & \makecell{Latency\\$\downarrow$} \\
\midrule
M0: Base direct & 77.6 & 66.5 & 0.489 & 0.158 & 0.058 & 2.2 & $1.0\times$ \\
M1: Explicit think & 82.3 & 75.8 & 0.377 & 0.120 & 0.045 & 1.3 & $6.0\times$ \\
M2: Preference SFT & 78.9 & 69.2 & 0.439 & 0.142 & 0.051 & 1.0 & $1.0\times$ \\
M3: Decision--reason & 79.9 & 71.2 & 0.409 & 0.133 & 0.046 & 0.8 & $1.0\times$ \\
M4: Hard KD & 80.4 & 72.3 & 0.501 & 0.154 & 0.062 & 1.6 & $1.0\times$ \\
M5: Soft KD & 81.7 & 74.8 & 0.385 & 0.122 & 0.041 & 0.6 & $1.0\times$ \\
\bottomrule
\end{tabular}
\end{table}

%% file: results_reward/hard_subsets.tex
\begin{table}[!htbp]
\centering\footnotesize
\caption{\textbf{Difficult-pair accuracy (\%).} Scores are shown for five error categories and the reported overall.}
\label{tab:reward-hard}
\setlength{\tabcolsep}{4pt}
\begin{tabular}{@{}lrrrr@{}}
\toprule
Category & M0 & M1 & M4 & M5 \\
\midrule
Wrong target or ordinal & 70.0 & 78.3 & 75.8 & 78.3 \\
Near-miss text & 60.0 & 72.5 & 68.3 & 70.8 \\
Preservation damage & 69.2 & 76.7 & 73.3 & 78.3 \\
Coupled-edit omission & 61.7 & 73.3 & 68.3 & 70.8 \\
Visually good, task wrong & 71.7 & 78.3 & 75.8 & 75.8 \\
Reported overall & 66.5 & 75.8 & 72.3 & 74.8 \\
\bottomrule
\end{tabular}
\end{table}

%% file: results_reward/external.tex
\begin{table}[!htbp]
\centering\footnotesize
\caption{\textbf{External RM evaluation.} MMRB2 editing pair accuracy and EditReward-Bench $K=2,3,4$ ranking scores are reported separately (\%).}
\label{tab:reward-external}
\setlength{\tabcolsep}{4pt}
\begin{tabular}{@{}lrrrr@{}}
\toprule
RM & \makecell{MMRB2 edit\\pair acc. $\uparrow$} & \makecell{EditReward\\$K=2$ $\uparrow$} & \makecell{EditReward\\$K=3$ $\uparrow$} & \makecell{EditReward\\$K=4$ $\uparrow$} \\
\midrule
M0 & 66.2 & 57.2 & 35.9 & 10.8 \\
M1 & 71.8 & 61.8 & 44.9 & 16.6 \\
M2 & 67.1 & 58.3 & 37.8 & 11.9 \\
M3 & 68.6 & 59.6 & 40.4 & 13.3 \\
M4 & 69.3 & 60.1 & 41.5 & 13.8 \\
M5 & 70.6 & 61.0 & 43.5 & 15.8 \\
\bottomrule
\end{tabular}
\end{table}

%% file: results_reward/downstream.tex
\begin{table}[!htbp]
\centering\footnotesize
\caption{\textbf{Reward $\times$ gate comparisons on Wan-Image.} D0--D3 form the $2\times2$ comparison; D4 uses explicit thinking. D0--D4 use equal update steps and rollout budgets. All metrics are percentages.}
\label{tab:reward-downstream}
\setlength{\tabcolsep}{4pt}
\begin{tabular}{@{}llrrr@{}}
\toprule
ID / RM & Gates & CCR $\uparrow$ & AC $\uparrow$ & Pres. $\uparrow$ \\
\midrule
D0 / M0 & Off & 69.08 & 88.91 & 91.35 \\
D1 / M0 & On & 74.25 & 91.28 & 93.12 \\
D2 / M5 & Off & 70.92 & 89.72 & 91.82 \\
D3 / M5 & On & 77.08 & 92.31 & 93.86 \\
D4 / M1 & On & 77.50 & 92.42 & 93.73 \\
\bottomrule
\end{tabular}
\end{table}

%% file: sections/human_evaluation.tex
\subsection{Human evaluation of final edits}
\label{sec:human-final}
We evaluate final edits on 200 CE-Bench tasks, allocated 50/40/40/40/30 across
categories A--E.
\Cref{tab:human-main} reports CCR, automatic--human bias, and agreement on
each method's human-decidable subset, together with false acceptance/rejection
and CCR bounds over all 200 tasks.

\input{results_analysis/human_main}

\paragraph{Matched gains and scoring bias.}
On 194 tasks with both human labels resolved, full improves human CCR over
ungated from 69.07\% to 75.77\%, a 6.70-point gain
[1.03, 12.37] (\cref{tab:human-paired}). The automatic gain is 6.19 points;
automatic minus human gain is $-0.52$ points [$-3.61$, 2.58].

\input{results_analysis/human_paired}

\input{figures/human_preference_in_paper}

\paragraph{Dimension-specific preference.}
On the same tasks, instruction preference favors full against both ungated
and direct, and preservation preference favors full against direct
(\cref{tab:human-preferences}). Visual-quality intervals in both comparisons
and the preservation interval against ungated span zero.
Net preference is $100(N_{\rm win}-N_{\rm loss})/N_{\rm decidable}$, including
ties. \Cref{fig:human-preference} shows wins, ties, and losses against ungated.

\input{results_analysis/human_preferences}

\paragraph{Confidence intervals.}
Percentile 95\% intervals use 50,000 category-stratified task bootstrap
resamples, preserving paired labels and each analysis's decidable counts.

%% file: results_analysis/human_main.tex
\begin{table}[!htbp]
\centering\footnotesize\setlength{\tabcolsep}{3.5pt}
\caption{\textbf{Final-output human evaluation on 200 CE-Bench tasks.}
$N$ is the number of human-decidable outputs per method. CCR, agreement, false
acceptance (FA), and false rejection (FR) are percentages; bias is automatic
minus human CCR in points. FA conditions on human failures and FR on human
successes. Identification bounds assign unresolved labels to failure or
success over all 200 tasks.}
\label{tab:human-main}
\begin{tabular}{@{}lrrrrrrrr@{}}
\toprule
Method & $N$ & \makecell{Human\\CCR} & \makecell{Auto.\\CCR} & Bias & Agree. & FA & FR & \makecell{Human CCR\\bounds} \\
\midrule
Direct & 195 & 53.33 & 54.36 & +1.03 & 88.72 & 13.19 & 9.62 & [52.0, 54.5] \\
Ungated & 196 & 69.39 & 70.92 & +1.53 & 91.33 & 16.67 & 5.15 & [68.0, 70.0] \\
Full & 197 & 76.14 & 76.65 & +0.51 & 93.40 & 14.89 & 4.00 & [75.0, 76.5] \\
\bottomrule
\end{tabular}
\end{table}

%% file: results_analysis/human_paired.tex
\begin{table}[!htbp]
\centering\small\setlength{\tabcolsep}{4pt}
\caption{\textbf{Full versus ungated on 194 common tasks.} Point estimates and 95\% percentile intervals are in percentage points. Intervals use 50,000 paired task bootstrap resamples within categories, preserving the observed category counts of this common decidable set.}
\label{tab:human-paired}
\begin{tabular}{@{}lrr@{}}
\toprule
Quantity & Estimate & 95\% interval \\
\midrule
Human gain $\Delta_H$ & +6.70 & [+1.03, +12.37] \\
Automatic gain $\Delta_A$ & +6.19 & [+1.55, +10.82] \\
Gain bias $\Delta_A-\Delta_H$ & -0.52 & [-3.61, +2.58] \\
\bottomrule
\end{tabular}
\end{table}

%% file: figures/human_preference_in_paper.tex
\begin{figure}[!htbp]
\centering
\includegraphics[width=\linewidth]{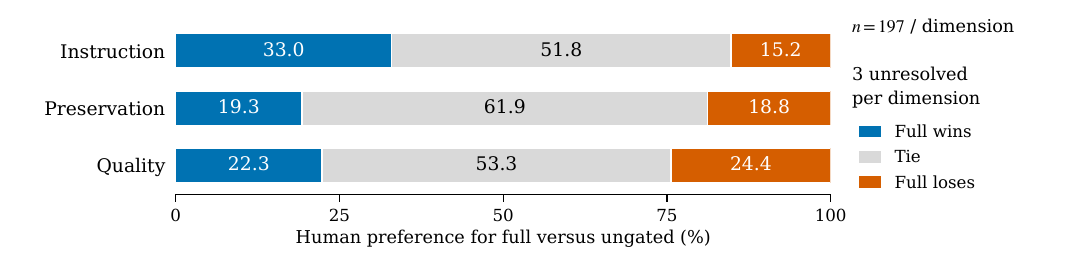}
\caption{\textbf{Human preferences for full versus ungated.} Each dimension has
197 decidable task pairs and three unresolved pairs out of 200. Wins, ties,
and losses retain their separate categories; interval estimates for net
preference appear in \cref{tab:human-preferences}.}
\label{fig:human-preference}
\end{figure}

%% file: results_analysis/human_preferences.tex
\begin{table}[!htbp]
\centering\footnotesize\setlength{\tabcolsep}{4pt}
\caption{\textbf{Human preference for full.} W/T/L are wins/ties/losses; Unres. counts unresolved judgments out of 200 task pairs per comparison. Net preference is 100(W-L)/(W+T+L), in points. The 95\% intervals use category-stratified task bootstrap; ties remain in the denominator.}
\label{tab:human-preferences}
\begin{tabular}{@{}llrrrr@{}}
\toprule
Full comparison & Dimension & W/T/L & Unres. & Net & 95\% interval \\
\midrule
vs. ungated & Instruction & 65/102/30 & 3 & +17.77 & [+8.63, +26.90] \\
vs. ungated & Preservation & 38/122/37 & 3 & +0.51 & [-8.12, +9.14] \\
vs. ungated & Quality & 44/105/48 & 3 & -2.03 & [-11.68, +7.11] \\
vs. direct & Instruction & 89/83/25 & 3 & +32.49 & [+22.84, +42.13] \\
vs. direct & Preservation & 70/94/33 & 3 & +18.78 & [+9.14, +28.43] \\
vs. direct & Quality & 61/87/48 & 4 & +6.63 & [-3.57, +16.84] \\
\bottomrule
\end{tabular}
\end{table}

%% file: results_t2i/generalization.tex
\begin{table}[!htbp]
\centering\small\setlength{\tabcolsep}{3pt}
\caption{\textbf{Text-to-image adaptation after retraining.} Metrics follow
\cref{sec:t2i-results}; dark/light pink mark the best/second-best values across
all six systems, including ties.}
\label{tab:t2i-generalization}
\begin{tabularx}{\linewidth}{@{}Xrrrrr@{}}
\toprule
Model & \makecell{GenEval2\\AM $\uparrow$} & \makecell{GenEval2\\GM $\uparrow$} & \makecell{R2I\\$\times100$ $\uparrow$} & \makecell{WISE-V\\$\times100$ $\uparrow$} & DPG $\uparrow$ \\
\midrule
BAGEL + native CoT & 70.9 & 23.1 & 48.0 & 62.8 & 85.1 \\
BAGEL + DuplexBridge-T2I & 77.8 & 31.5 & 56.0 & \BestScore{70.5} & 86.8 \\
\midrule
Qwen-Image & 80.8 & 33.8 & 52.0 & 51.0 & 88.3 \\
Qwen-Image + DuplexBridge-T2I & \BestScore{85.0} & \BestScore{42.0} & \BestScore{61.0} & \SecondScore{66.0} & \BestScore{89.7} \\
\midrule
Wan-Image-7B & 79.4 & 32.2 & 51.0 & 54.0 & 87.7 \\
Wan-Image-7B + DuplexBridge-T2I & \SecondScore{84.2} & \SecondScore{40.6} & \SecondScore{59.5} & 64.5 & \SecondScore{89.3} \\
\bottomrule
\end{tabularx}
\end{table}

%% file: sections/7_qualitative.tex
\section{Qualitative Comparisons}
\label{sec:qualitative}

\begin{minipage}{\linewidth}
\input{figures/rise_qualitative_in_paper}

\par\vspace{12pt}
\input{figures/kris_qualitative_in_paper}
\end{minipage}

\begin{figure}[p]
\centering
\includegraphics[width=\linewidth,height=0.88\textheight,keepaspectratio]{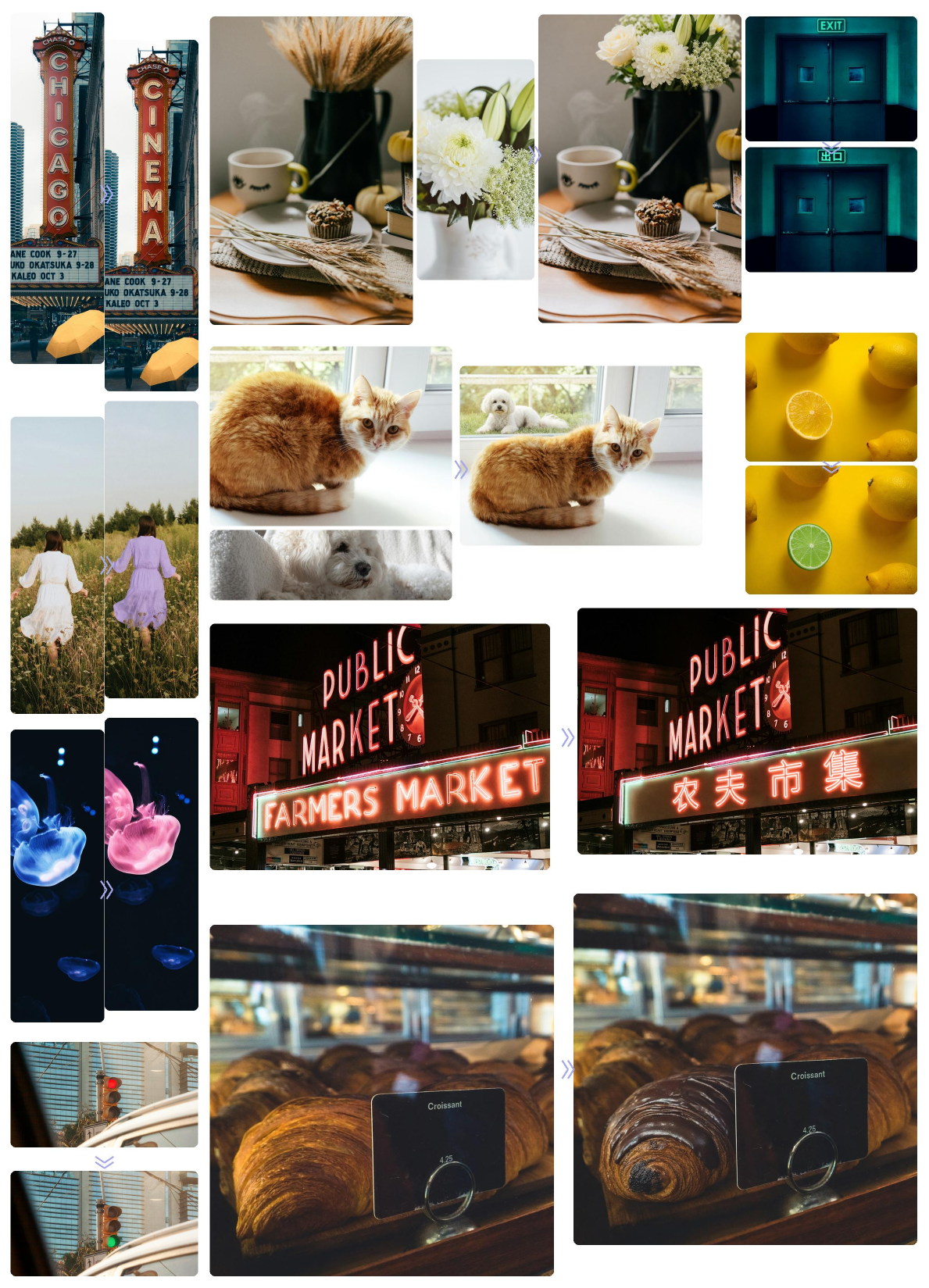}
\caption{\textbf{Additional CE-Bench showcase.} Arrows indicate the transition
from source images to edited outputs; auxiliary images provide visual references
for flower replacement and dog insertion. Examples include sign-text replacement
and translation, localized color changes, a traffic-light state change, and
object appearance edits. The paired images show the edited regions alongside
the surrounding scene.}
\label{fig:ce-showcase}
\end{figure}
\FloatBarrier

\paragraph{RISEBench.}
In \cref{fig:rise-qualitative}, DuplexBridge draws a connected blue maze route
through the white corridor between the red and green endpoints. The untied
rope and opaque egg whites show physical state changes. In the equivalence
example, DuplexBridge replaces the question mark with two green cylinders,
consistent with the displayed relation and the original task's request for
the correct number of cylinders.

\paragraph{KRIS-Bench.}
The examples in \cref{fig:kris-qualitative} illustrate coupled requirements:
a balance responds to added load, oil separates above water, and three cups
reach a common size. Reference-guided garment replacement preserves the white
trousers. These examples accompany the quantitative public-benchmark results
in \cref{tab:public-benchmarks-main}.

%% file: figures/rise_qualitative_in_paper.tex
\begin{minipage}{\linewidth}
\captionsetup{type=figure,skip=5pt}
\centering
\includegraphics[width=\linewidth]{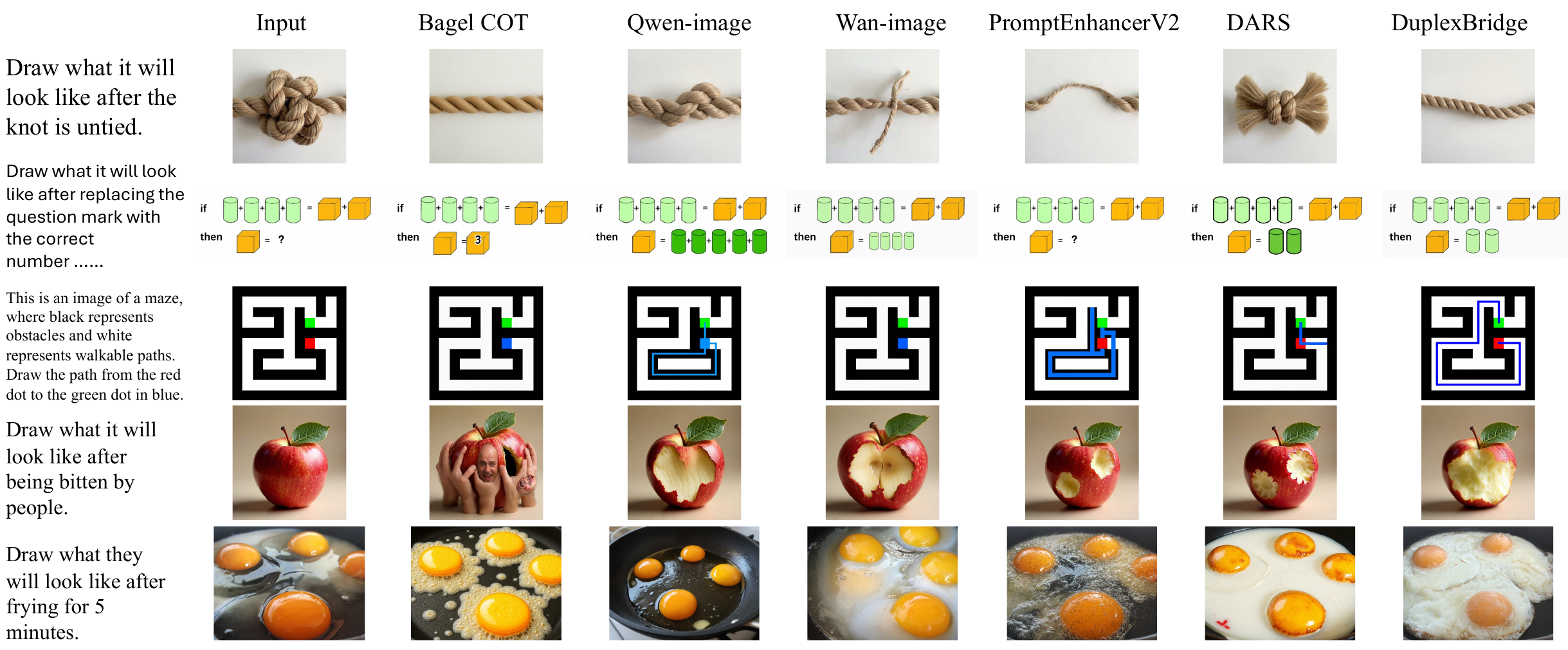}
\caption{\textbf{Qualitative comparison on RISEBench.} Rows cover
untying a knot, a visual equivalence relation, maze navigation, a bitten apple,
and cooking progression. The maze requires a connected path through the white
corridor; the physical transformations also require recognizable source content.}
\label{fig:rise-qualitative}
\end{minipage}

%% file: figures/kris_qualitative_in_paper.tex
\begin{minipage}{\linewidth}
\captionsetup{type=figure,skip=5pt}
\centering
\includegraphics[width=\linewidth]{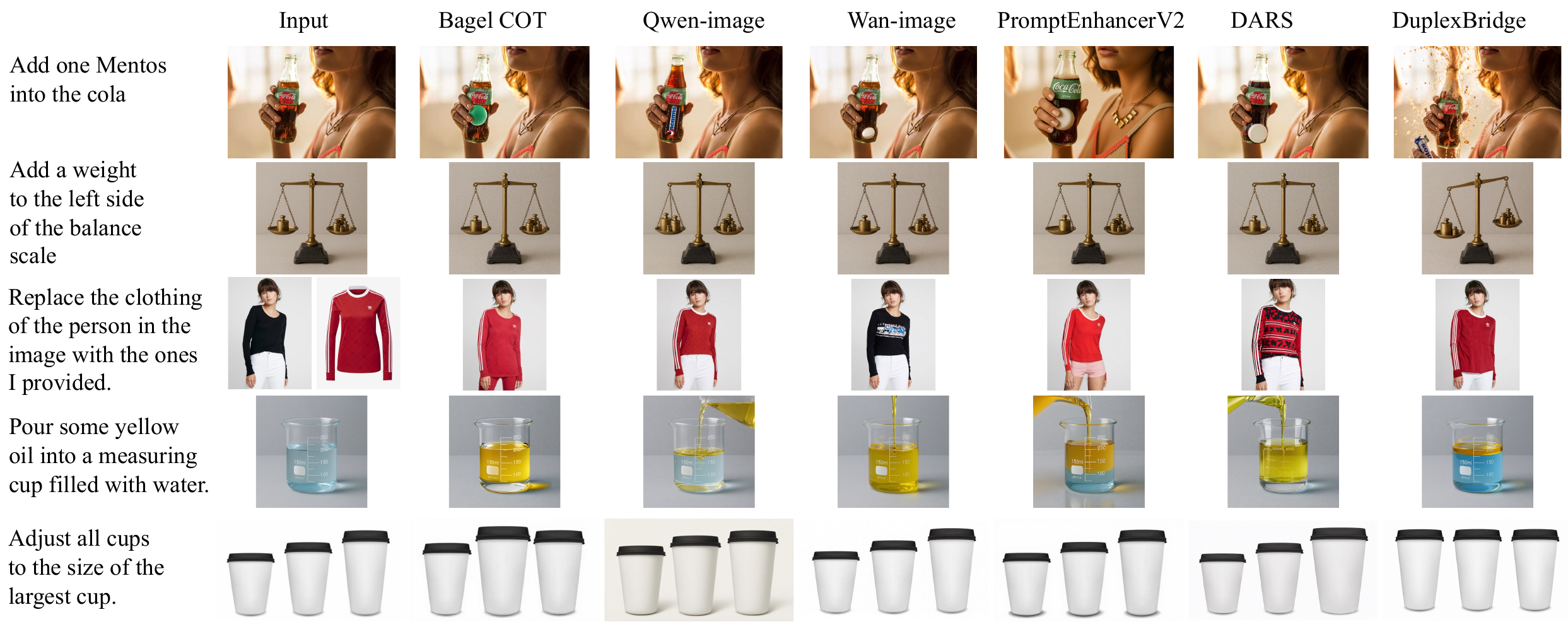}
\caption{\textbf{Qualitative comparison on KRIS-Bench.} The tasks combine requested
changes with physical consequences, reference fidelity, and preservation.
Rows show Mentos--cola interaction, added load, garment replacement,
oil--water separation, and cup-size equalization.}
\label{fig:kris-qualitative}
\end{minipage}

%% file: sections/8_discussion.tex
\section{Limitations}
\label{sec:discussion}
\label{sec:limitations}
Our experiments demonstrate substantial improvements in image editing and
generation, but DuplexBridge has not yet been evaluated on more complex video
generation and editing tasks. Extending the method to the video domain is a
promising direction for future research.

%% file: sections/7_conclusion.tex
\section{Conclusion}
\label{sec:conclusion}
DuplexBridge connects a unified model's own understanding and generation
through a checkable editing condition. Local supervision distinguishes task
fidelity from received-condition execution, while reciprocal trust gates
select additional global feedback and retain valid local learning.
Across three configurations, complete compliance improves over direct editors;
controlled Wan-Image comparisons support the supervision references and
semantic selection, with gains also supported by human evaluation.
These findings support separating local responsibilities when jointly
adapting understanding and generation.
\par
\label{sec:body-end}